\documentclass[final,3p,times,twocolumn,authoryear]{elsarticle}     

\usepackage{amssymb}
\usepackage{amsmath}

\usepackage{float}          
\usepackage{booktabs}       
\usepackage{subcaption}     
\usepackage{url}            
\journal{ISPRS Open Journal of Photogrammetry and Remote Sensing}

\begin{document}

\begin{frontmatter}



\title{Deep Learning-based Robust Autonomous Navigation of Aerial Robots in Dense Forests} 


\author[inst1]{Guglielmo Del Col \corref{correspondingAuthor}}
\ead{guglielmo.delcol@nls.fi}
\affiliation[inst1]{organization={Department of Remote Sensing and Photogrammetry, Finnish Geospatial Research Institute FGI, The National Land Survey of Finland},
            addressline={Vuorimiehentie 5}, 
            city={Espoo},
            postcode={02150}, 
            country={Finland}}

\author[inst1]{V\"{a}in\"{o} Karjalainen}
\author[inst1]{Teemu Hakala}
\author[inst1]{Yibo Zhang}
\author[inst1]{Eija Honkavaara}

\cortext[correspondingAuthor]{Corresponding author}

\begin{abstract}
Autonomous aerial navigation in dense natural environments remains challenging due to limited visibility, thin and irregular obstacles, GNSS-denied operation, and frequent perceptual degradation. This work presents an improved deep learning–based navigation framework that integrates semantically enhanced depth encoding with neural motion-primitive evaluation for robust flight in cluttered forests. Several modules were incorporated on top of the original sevae-ORACLE algorithm, to address limitations observed during real-world deployment, including lateral control for sharper maneuvering, a temporal-consistency mechanism to suppress oscillatory planning decisions, a stereo-based VIO solution for drift-resilient state estimation, and a supervisory safety layer that filters unsafe actions in real time. A depth-refinement stage was additionally included to improve the representation of thin branches and reduce stereo noise, while GPU optimization increased onboard inference throughput from 4 Hz to 10 Hz.

To contextualize system performance, a comparative evaluation was conducted against several existing learning-based navigation methods tested under identical environmental conditions and hardware constraints. The proposed approach demonstrated consistently higher success rates, more stable trajectories, and improved collision avoidance, particularly in highly cluttered forest settings.

The system was deployed on a custom quadrotor in three boreal forest environments. In moderate-clutter and dense environments, it achieved fully autonomous completion in all 15/15 flights, while in a highly dense underbrush setting it succeeded in 12/15 trials, representing a substantial improvement over the previous baseline, which completed only 5/15 comparable missions. These findings demonstrate that the proposed system provides a measurable improvement in reliability and safety over existing navigation methods in complex natural environments.
\end{abstract}


\end{frontmatter}

\section{Introduction}\label{Introduction}
\sloppy

Autonomous navigation in cluttered natural environments such as forests remains a fundamental challenge in robotics. Under forest canopies, Global Navigation Satellite Systems (GNSS) often become unreliable, mandating fully onboard navigation pipelines based solely on visual, inertial, or proximity sensors. In addition, tree trunks, dense foliage, and thin branches create visually irregular scenes with pervasive occlusions, abrupt depth discontinuities, and highly variable illumination. Under forest canopies, Global Navigation Satellite Systems (GNSS) often become unreliable, mandating fully onboard navigation pipelines based solely on visual, inertial, or proximity sensors. These conditions make accurate motion estimation, reliable obstacle detection, and robust collision-free planning exceedingly demanding, particularly when small-scale vegetation elements compromise perception fidelity.

Beyond the robotics challenge itself, dependable under-canopy autonomy is increasingly recognized as an enabler for modern forest measurement and ecological monitoring. Recent work has shown that lightweight under-canopy UAVs can acquire close-range photogrammetric and LiDAR data for individual tree structural estimation, including canopy height and stem volume measurements comparable to above-canopy surveys \citep{Hyyppa2021Under}, and can be integrated with above-canopy sensing to reconstruct detailed forest inventories including stem attributes \citep{wang2021seamless}.
In particular, \citet{Karjalainen2025towards} demonstrated that compact under-canopy robotic drones can be used to capture image data suitable for forest inventory even in highly cluttered GNSS-denied environments. These developments underscore the growing demand for reliable autonomous flight beneath forest canopies, yet the navigation systems required to support such applications are still far from mature.

Classical Simultaneous Localization and Mapping (SLAM) techniques, while effective in structured indoor or urban settings, often degrade in dense natural environments. Pose-graph SLAM approaches frequently accumulate drift or produce inconsistent maps due to perceptual aliasing in repetitive forest scenes \citep{Cadena2016PastAge, Ebadi2022PresentEnvironments, Sandstrom2023UncLe-SLAM:SLAM}. Volumetric mapping solutions based on uniform voxel grids, Truncated Signed Distance Function (TSDF) reconstructions, or distance fields (e.g., OctoMap, TSDF-based planners, and VDB-based approaches) depend on dense and reliable depth or pose estimates \citep{Hornung2013OctoMap:Octrees, Oleynikova2017Voxblox:Planning, Millane2024Nvblox:Mapping}. Unfortunately, stereo and RGB-D sensors often fail in cluttered undergrowth, where thin branches, self-similar textures, and low light yield missing or noisy returns \citep{Oleynikova2017Voxblox:Planning, Barry2018High-speedStereo, Matthies2014StereoSpace}. LiDAR-based mapping can mitigate some of these limitations, but payload, energy consumption, and computational demands typically exceed the constraints of lightweight UAV platforms \citep{Zhang2024Uav-borneApplications}.

Recognizing these challenges, research has increasingly shifted toward learning-based, map-free navigation paradigms that attempt to bypass explicit mapping and plan directly from raw sensor data. Reactive pipelines using compressed representations, such as kd-trees \citep{Florence2020IntegratedMaps}, circular buffers \citep{Usenko2017Real-timeBuffer}, or disparity-based processing from depth cameras \citep{Matthies2014StereoSpace}, enable low-latency control decisions. More advanced deep learning frameworks have enabled real-time inference of control-relevant quantities, such as traversability or collision risk, directly from raw images \citep{Gandhi2017LearningCrashing, Loquercio2018DroNet:Driving}. Reinforcement learning with domain randomization has further improved sim-to-real robustness \citep{Sadeghi2016CAD2RL:Image, LoquercioLearningWild}. Model-free systems such as BADGR \citep{Kahn2021BADGR:System} or LaND \citep{Kahn2021LaND:Disengagements} predict collision likelihood or traversability from onboard sensory streams, enabling exploration and navigation in previously unseen cluttered environments.

However, many of these learning-based solutions simplify real-world dynamics by ignoring momentum, actuator delays, or the coupling between translational and rotational motion, or by assuming nearly complete environmental observability \citep{Pfeiffer2018ReinforcedDemonstrations}. Such simplifications break down under forest conditions, where motion blur, missing depth returns, and partial sensor failure are common due to occlusion and complex geometry \citep{Nguyen2022MotionPrediction, Khattak2020ComplementaryEnvironments}. To address these issues, recent works have incorporated probabilistic reasoning or semantic scene structure. Examples include uncertainty-aware collision prediction under partial observability \citep{Nguyen2024Uncertainty-awareNetworks} and semantic latent-space encoding that preserves fine structural detail, including thin branches, within low-dimensional representations \citep{Kulkarni2023AerialRobots}.
In particular, Nguyen et al. \citep{Nguyen2024Uncertainty-awareNetworks} demonstrated the practical feasibility of their ORACLE framework through a real-world flight in a dense Finnish forest in Evo, successfully traversing approximately 60 meters of cluttered vegetation with a single autonomous run.

Beyond purely learning-based approaches, recent studies have attempted to deploy autonomous UAVs under dense forest canopy using classical or hybrid methods. For example, \citet{Karjalainen2023ARESULTS} evaluated a prototype of a robotic under-canopy drone that combined VINS-Fusion \citep{Qin2018VINS-Mono:Estimator, qin2019local} for pose estimation, stereo-depth mapping, and a reactive trajectory planner EGO-planner-v2 \citep{Zhou2022SwarmWild}. Their real-world tests in boreal forest plots showed promising performance in light to moderate forest density, but also revealed substantial difficulties in reliably detecting and avoiding small branches and understory vegetation in dense stands \citep{Karjalainen2023ARESULTS}. Similarly, \citet{karhunen2025fieldevaluationoptimizationlightweight} evaluated a LiDAR-based autonomous flying system using a Livox Mid-360 on a custom quadrotor platform, integrating LTA-OM SLAM \citep{zou2024lta} for LiDAR–inertial odometry and an IPC path planner. Their extensive testing across forest plots of varying density demonstrated reasonable performance in medium-density stands, 12 of 15 successful flights, but substantially lower performance in dense forest environments, 5 of 15 successful flights, at 2 m/s \citep{karhunen2025fieldevaluationoptimizationlightweight}.

These results confirm that although recent UAV-based solutions have advanced forest autonomy, significant gaps remain, particularly in dense canopy conditions where sensors struggle to perceive fine obstacles and planning systems must respond rapidly under uncertainty.

This work builds on previous contributions \citep{DelColGuglielmo2024Autonomous, DelColGuglielmo2024RefiningLearning}, which introduced a simulation-trained pipeline that combines a semantically enhanced depth encoder \citep{Kulkarni2023Semantically-enhancedRobots} with a neural collision prediction network \citep{Nguyen2024Uncertainty-awareNetworks}. Initial field trials demonstrated feasibility, but also revealed limitations related to lateral control authority, planning oscillations, sensitivity to stereo noise in thin structures, and VIO drift, using a RealSense T265 camera, under repetitive forest textures.

Significant enhancements have been introduced to mitigate these limitations and to improve robustness in real-world navigation. These include (I) the extension of the action space to include lateral velocity commands for sharper curved trajectories, (II) a temporal-consistency mechanism to suppress oscillatory decisions in cluttered foliage, (III) improved pose estimation through open-source stereo VIO algorithm, VINS-Fusion \citep{Qin2018VINS-Mono:Estimator, qin2019local}, for more drift-resilient odometry, (IV) a real-time supervisory safety layer that filters unsafe motion primitives, (V) a depth-refinement module to suppress stereo noise and better represent fine obstacles, and (VI) porting the collision prediction network to TensorRT for efficient onboard inference at 10 Hz.

A comparative evaluation is also planned between the enhanced method, the LiDAR-based system of \citet{karhunen2025fieldevaluationoptimizationlightweight}, the stereo–VIO mapping prototype of \citet{Karjalainen2023ARESULTS}, and the original SEVAE-ORACLE baseline by \citet{Nguyen2024Uncertainty-awareNetworks}. This comparison will highlight trade-offs between sensing modalities and real-world performance under different forest densities.


Field experiments were conducted in three boreal forest environments of increasing complexity to evaluate the robustness and generalization capabilities of the proposed navigation system. The aim of these real-world tests is to demonstrate that the method can handle diverse vegetation structures, from moderately cluttered stands to highly dense underbrush, while maintaining reliable obstacle avoidance and stable flight. This work seeks to advance deployable aerial autonomy in complex natural environments and to support future applications in forestry surveying, environmental monitoring, and search-and-rescue operations in vegetation-rich forests.


The remainder of this study is structured as follows. Section 2 describes the methods, algorithms, and hardware utilized in this study, as well as the field-experiments carried out. The experimental results are presented in Section 3. Section 4 discusses the results and potential improvements followed by the conclusion of the study in Section 5.

The full implementation of the proposed autonomous forest navigation system is available open-source in the DeFoP GitHub repository\footnote{\url{https://github.com/guglielmo610/DeFoP}}.

\section{Material and Methods} 

This section presents the hardware and software components of the autonomous drone navigation system and describes the end-to-end architecture developed for real-world deployment in cluttered forest environments. The system integrates semantically-aware perception, uncertainty-aware motion evaluation, and reactive control within a tightly coupled onboard processing pipeline. All modules operate within the Robot Operating System (ROS) \citep{QuigleyROS:System} environment and run synchronously at 30 Hz, apart for the collision prediction network that gives the velocity commands at a 10 Hz frequency. 

\subsection{Algorithm}

\begin{figure*}[t]
  \centering
  \includegraphics[width=0.85\textwidth]{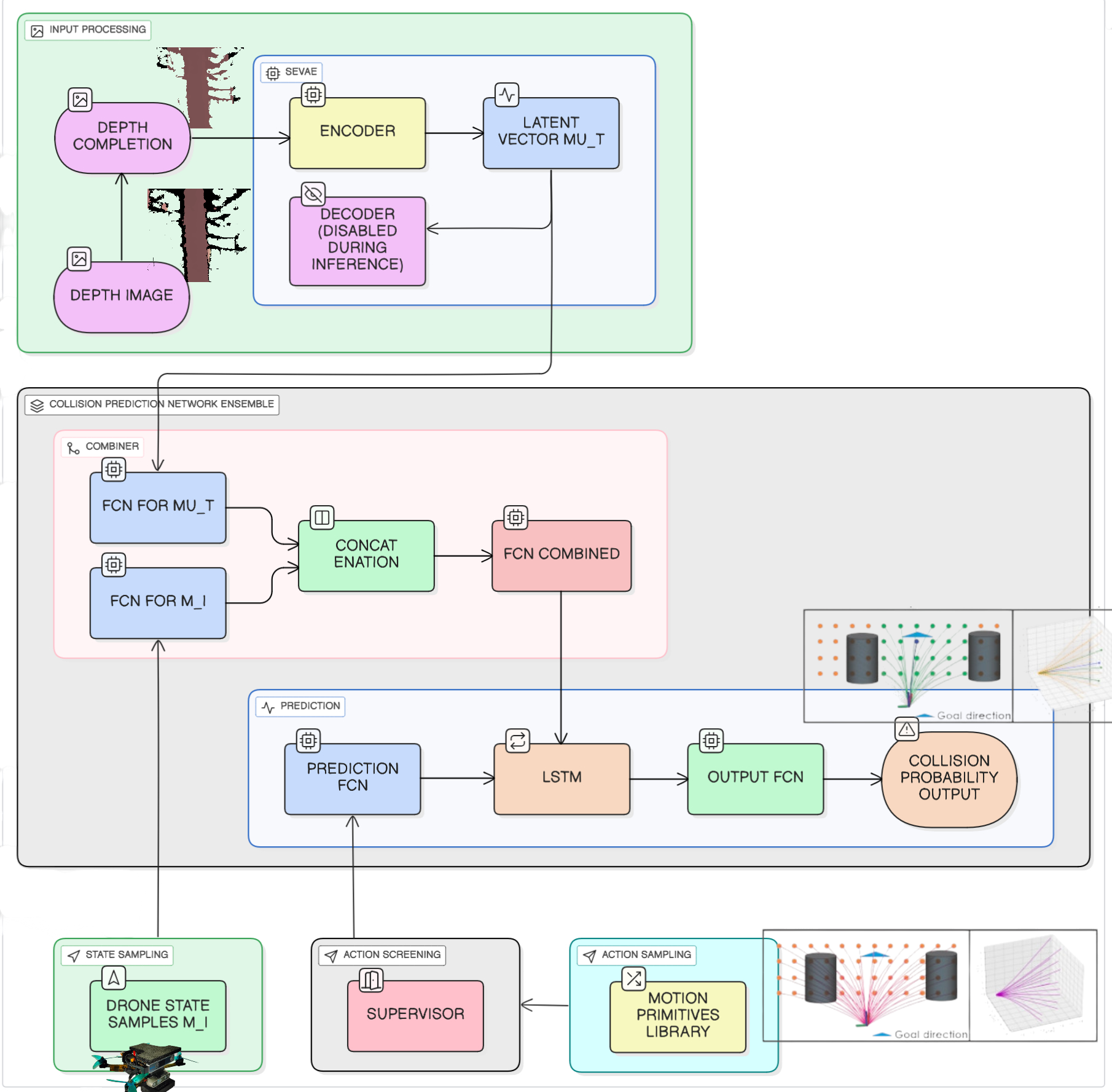}
  \caption{Navigation and planning pipeline.}
  \label{fig:pipeline}
\end{figure*}

This subsection describes the end-to-end navigation algorithm, including the perception pipeline, depth encoding, collision prediction and trajectory evaluation, safety supervision, and
state estimation. A schematic of the full navigation and planning pipeline is illustrated in Figure \ref{fig:pipeline}.

\subsubsection{System Architecture Overview}

The system comprises four main subsystems: (I) a perception module based on a convolutional autoencoder for depth compression and semantic enhancement, (II) a motion evaluation module employing a learned Collision Prediction Network (CPN), (III) a geometric safety supervisor module that enforces collision-free control, and (IV) a reactive control interface integrated with a PX4-based flight controller. 
The perception stack employs an Intel RealSense D435i (Intel, Santa Clara, California, U.S.A.) stereo camera to capture synchronized infrared and depth images at 270×480 resolution.

\begin{figure}[ht!]
    \centering

    \begin{subfigure}{0.48\columnwidth}
        \centering
        \includegraphics[width=\linewidth]{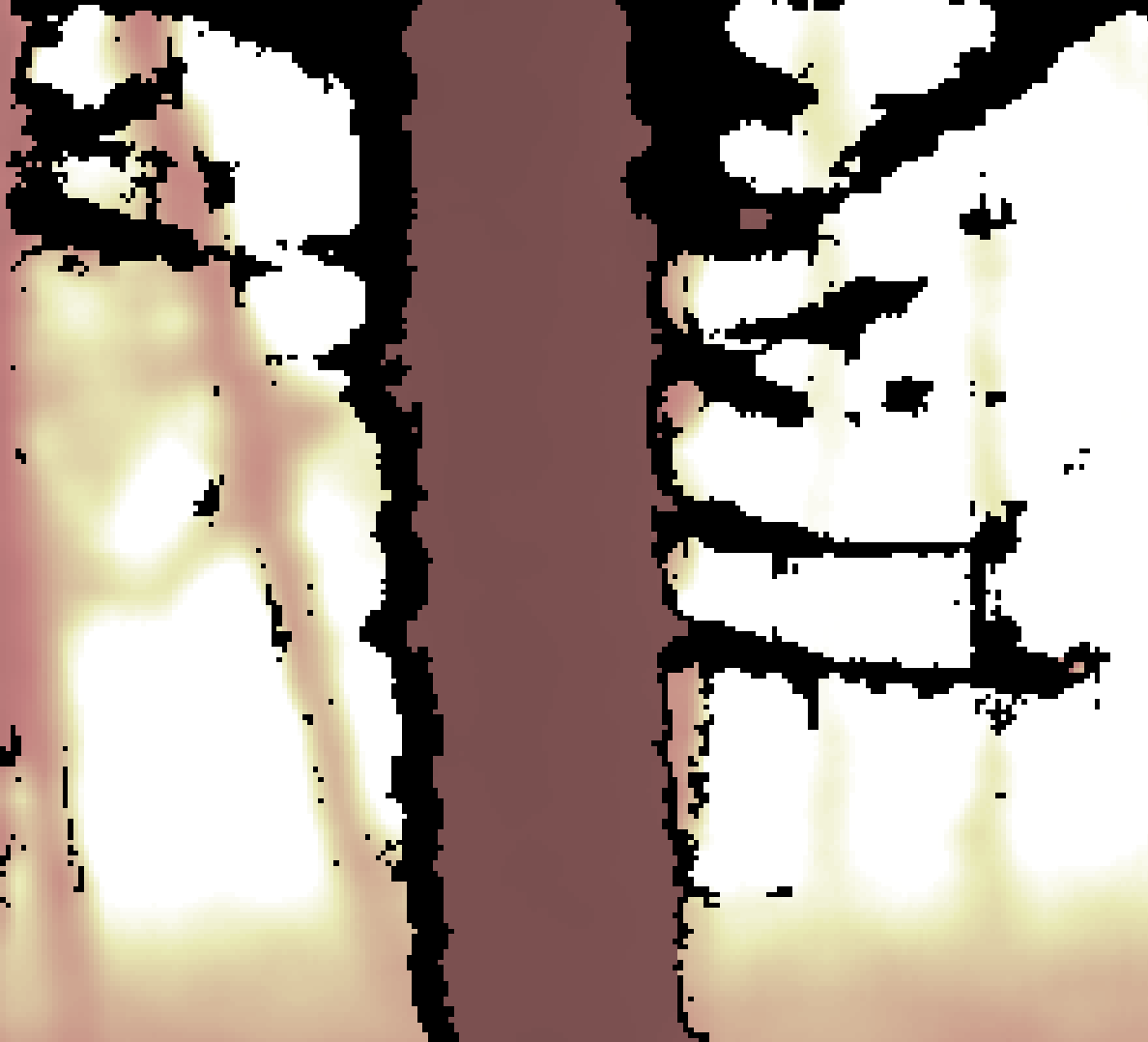}
        \caption{}
    \end{subfigure}
    \hfill
    \begin{subfigure}{0.48\columnwidth}
        \centering
        \includegraphics[width=\linewidth]{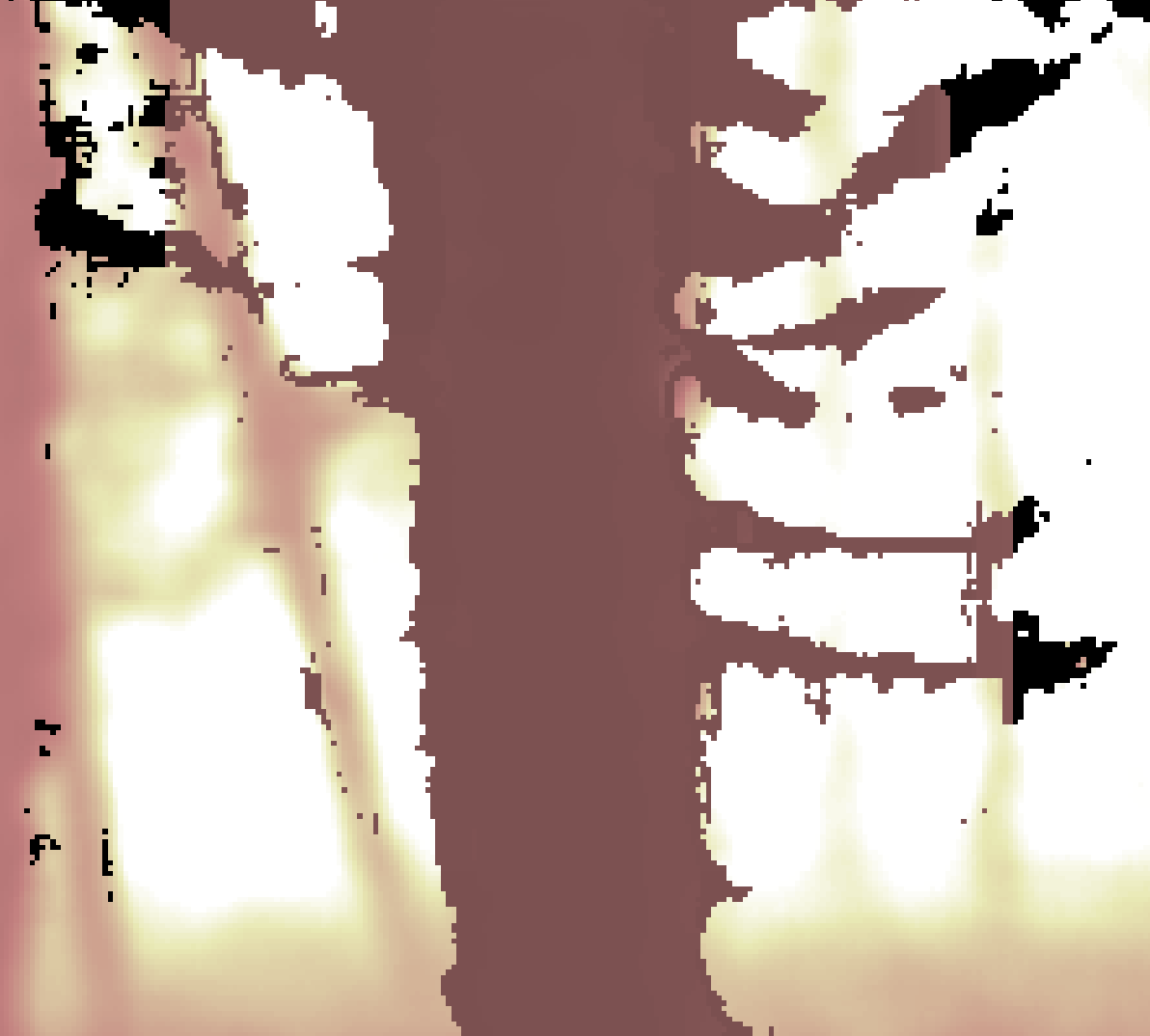}
        \caption{}
    \end{subfigure}

    \caption{Example illustrating the effect of depth-map refinement. (a) Raw depth image acquired from the Intel RealSense D435i. Colors encode the distance to obstacles using a brown-scale colormap, where darker tones indicate closer objects. Black pixels correspond to undefined depth measurements, which occur predominantly along object boundaries and thin structures, impairing accurate estimation of trunk thickness and branch presence. (b) Refined depth map after applying the proposed depth improver. Undefined pixels in the vicinity of obstacles are reconstructed, while regions far from obstacles remain undefined. This selective refinement improves obstacle representation while avoiding the introduction of spurious structures, thereby supporting safer trajectory selection.}
    \label{fig:depth_impr}
\end{figure}

Depth estimation is performed onboard the camera using an active stereo vision pipeline. Depth is computed by estimating pixel-wise disparity between the two synchronized global-shutter infrared images through a stereo matching process based on Semi-Global Matching (SGM) \citep{Hirschmueller2008Stereo}. An infrared pattern projector is used to enhance scene texture in low-feature regions, improving correspondence reliability under challenging lighting conditions. The resulting disparity maps are subsequently converted into metric depth measurements and streamed to the onboard computer for further processing.
Depth maps are preprocessed in real time to remove invalid pixels and refine noisy regions through a localized spatial propagation technique. This “depth improver” applies a kernel around regions where obstacles are detected closer than 2 m, filling undefined pixels by interpolating neighboring valid values. This operation enhances the continuity of thin branches and fine details that are often missing in raw stereo outputs. As an example of the depth improver in operation, Fig.~\ref{fig:depth_impr} shows how previously undefined pixels from the depth camera are almost completely recovered. The filtered depth maps are then forwarded to a 7-layer convolutional autoencoder \citep{Kulkarni2023Semantically-enhancedRobots} trained on synthetic data from RotorS \citep{Furrer2016rotorS} and Aerial Gym \citep{Kulkarni2023AerialRobots} environments, emphasizing semantically meaningful structural features such as thin branches and small vertical trunks. The encoder compresses each depth frame into a 128-dimensional latent representation that preserves spatial cues critical for obstacle detection and path safety estimation. 





\subsubsection{Collision Prediction and Planning Framework} 

The planning subsystem is structured around a Collision Prediction Network (CPN) inspired by the ORACLE architecture \citep{Nguyen2024Uncertainty-awareNetworks} and described in Section 2.1.4 . At each control cycle, the system generates a discrete set of 256 motion primitives defined by combinations of translational velocities (x-axis), vertical motion (z-axis) and yaw rates. These primitives represent potential short-horizon trajectories executable by the drone. Each candidate primitive is evaluated by the CPN, which receives as input the latent feature vector from the perception module and a partial state estimate of the drone, including linear velocities and orientation.

The CPN outputs a collision probability for each trajectory, and between the safest primitives, corresponding to the lowest predicted risk, the one that fits the most the final target direction and altitude is selected for execution. An enhancement over the prior system is the inclusion of a lateral velocity component (y-axis) in the control command set. Unlike the other motion primitives, this lateral velocity is not discretized but dynamically determined as a function of the commanded yaw rate. Specifically, it is set to the arctangent of the yaw command, allowing the drone to execute large, fast curves with improved responsiveness. This modification enables the planner to generate smoother and more agile trajectories, significantly enhancing obstacle avoidance performance in cluttered environments.

A heuristic-based planning stabilization mechanism addresses a commonly encountered issue during navigation in highly cluttered forests, frequent indecision at the planning level. This behavior is characterized by rapid oscillations in the selected yaw direction, resulting in reduced confidence and inefficient flight behavior. The system monitors the sign change frequency of yaw commands over a temporal window to detect this condition. When planning indecision is identified, a bias is applied to favor directional consistency, promoting smoother and more assertive trajectories. To further enhance navigation safety in dense clutter, the system is trained with an expanded collision margin from the original drone radius. By penalizing trajectories that pass close to obstacles during training, the CPN develops a conservative bias, producing flight paths that maintain greater clearance from branches and trunks. This modification substantially improves robustness in forest environments with abundant thin structures.

\subsubsection{Semantically Enhanced Autoencoder}

Effective autonomous navigation in cluttered forest environments requires accurate obstacle perception as well as compact and semantically meaningful scene representations that prioritize thin and sparse obstacles such as branches. The semantically-aware depth encoding framework proposed by \citet{Kulkarni2023Semantically-enhancedRobots} is adopted and extended, based on a convolutional autoencoder architecture designed to preserve fine geometric details in depth imagery. The autoencoder architecture consists of a symmetric encoder-decoder structure with seven convolutional and deconvolutional layers, respectively. 

The encoder sequentially compresses the spatial resolution while increasing the feature dimensionality, ultimately projecting each depth image into a 128-dimensional latent representation. This latent space serves as an input to the downstream collision prediction module and is optimized to retain critical scene geometry relevant to obstacle avoidance. The decoder, used only during training, mirrors the encoder to reconstruct the input image, facilitating a reconstruction loss that guides the preservation of visual content.

The model was pretrained with synthetic data generated by the Aerial Gym simulator \citep{Kulkarni2023AerialRobots}, which provides randomized tree and branch configurations. During fine-tuning, models of trunks and branches were selected from the Aerial Gym to closely resemble the structure and density characteristic of Finnish forests.

\begin{figure}[H]
    \centering

    \begin{subfigure}{\columnwidth}
        \centering
        \includegraphics[width=\columnwidth]{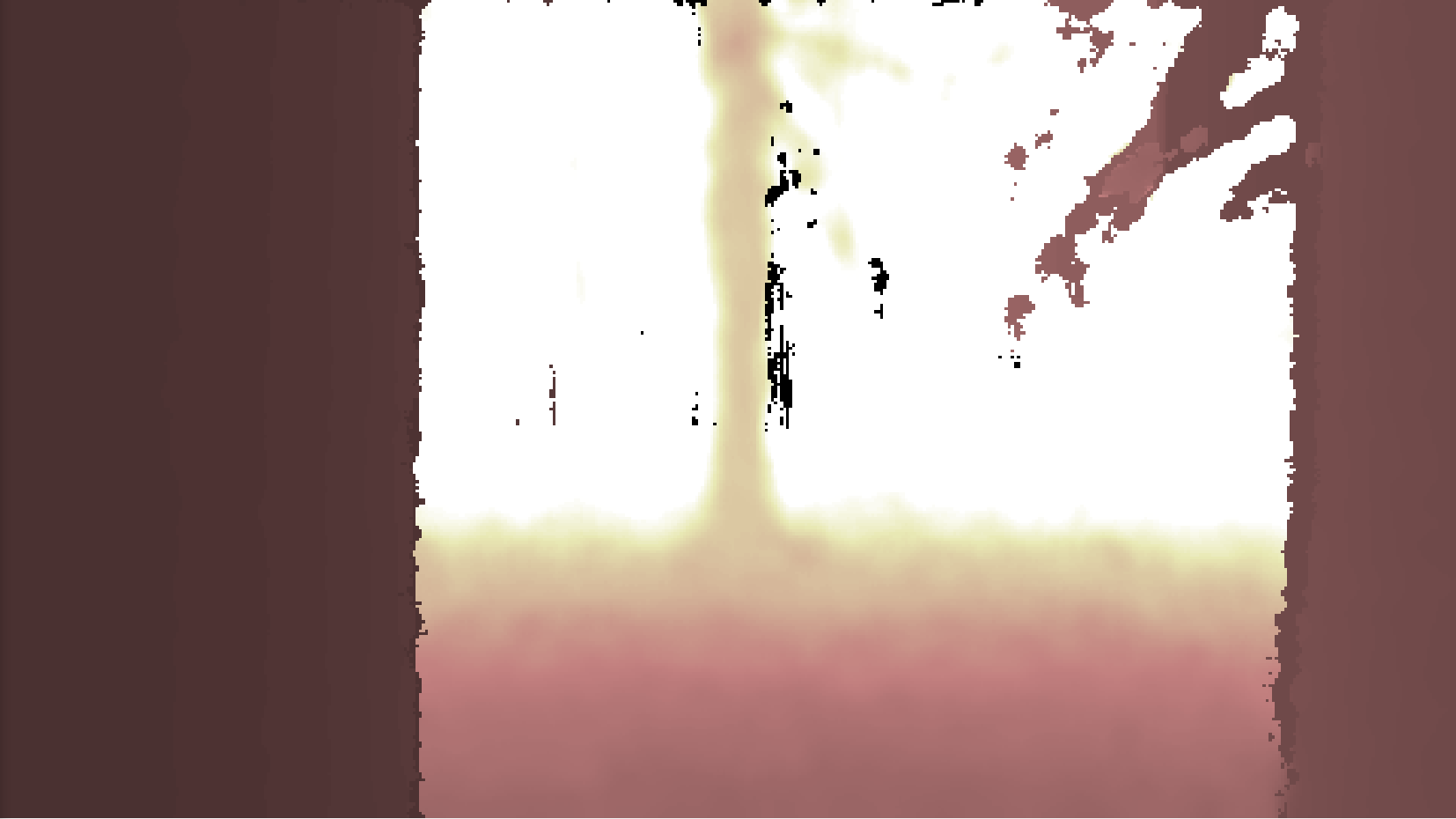}
        \caption{}
        \label{fig:depth_raw}
    \end{subfigure}

    \vspace{2mm}

    \begin{subfigure}{\columnwidth}
        \centering
        \includegraphics[width=\columnwidth]{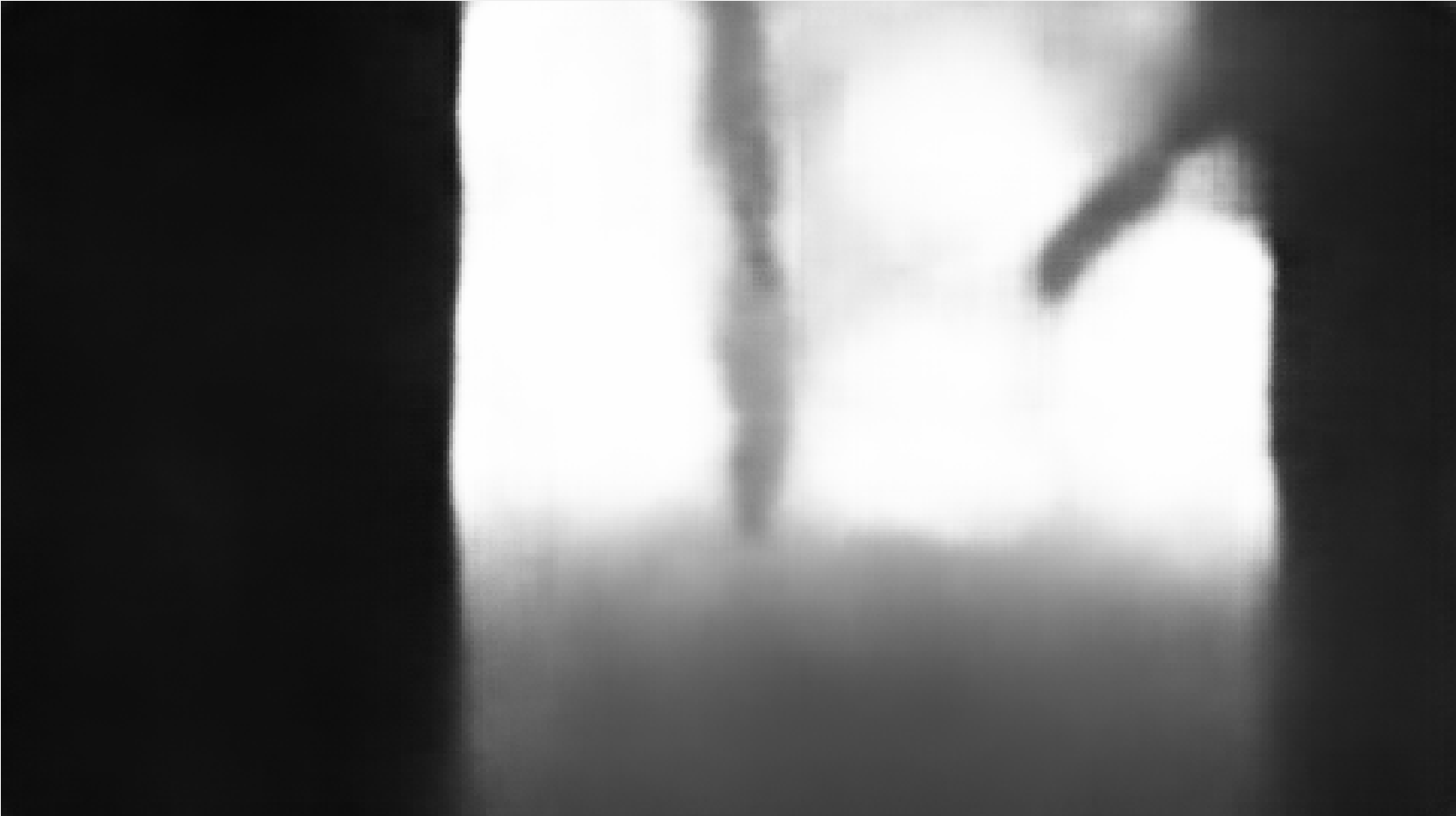}
        \caption{}
        \label{fig:depth_reconstructed}
    \end{subfigure}

    \caption{Example of autoencoder reconstruction. (a) Raw depth image. (b) Reconstructed depth image obtained after processing through the encoder–decoder convolutional layers. Notably, the semantically enhanced autoencoder is able to reconstruct the thin branch on the right side of the image almost completely, despite it being represented by only a small number of pixels in the input. This behavior is not typically observed in conventional autoencoders and highlights the effectiveness of the proposed semantic reconstruction strategy \citep{Kulkarni2023Semantically-enhancedRobots}.}
    \label{fig:reconstruction}
\end{figure}

Training was performed using a compound loss function that balances reconstruction accuracy and latent space regularization:  
\begin{equation} 
\mathcal{L}_{\text{total}} = \mathcal{L}_{\text{MSE}}^{\text{sem}} + \lambda \cdot \mathcal{L}_{\text{KLD}} 
\label{eq:total_loss} 
\end{equation} 
where $\mathcal{L}_{\text{MSE}}^{\text{sem}}$ represents the pixelwise mean squared error weighted by semantic importance, and $\mathcal{L}_{\text{KLD}}$ is the Kullback-Leibler divergence regularization term, encouraging a smooth latent distribution. Semantic weighting prioritizes visually sparse yet safety-critical structures such as branches, while invalid pixels, for example those missing due to sensor noise, are ignored through masking. Quantitative evaluations and qualitative reconstructions indicate substantial improvement in the preservation of thin obstacles and fine details of the scene. The latent embeddings demonstrate enhanced discriminability for thin, low-texture branches, even under varying lighting conditions or partial occlusions, scenarios in which conventional depth encoders often fail, as shown in Fig. \ref{fig:reconstruction}.





\subsubsection{Collision Prediction Network}

At the core of the autonomous navigation pipeline is the Collision Prediction Network (CPN), a data-driven decision-making module responsible for evaluating the safety of candidate motion trajectories based on current sensory observations. This module is adopted from the architecture proposed by \citet{Nguyen2022MotionPrediction, Nguyen2024Uncertainty-awareNetworks}, which demonstrated state-of-the-art performance in visually guided navigation under uncertainty.\\

In the current system, the original CPN architecture is retained with minor modifications, primarily in the data preprocessing and integration pipeline, to ensure compatibility with the latent representations produced by the semantically enhanced autoencoder and the realtime operating constraints of the onboard drone system. The CPN receives as input the compressed latent vector of the current depth image, extracted via the autoencoder described in subsection 2.1.3, in addition to a partial drone state vector comprising inertial velocity estimates and attitude information. These inputs jointly provide information about both the spatial structure of the environment and the dynamical state of the drone. The network predicts the likelihood of collision for a discrete set of predefined motion primitives.

Each primitive defines a short-horizon trajectory and is parameterized by a fixed forward velocity along the x-axis, a discrete vertical velocity sampled from eight quantized values along the z-axis, and a yaw rate selected from a set of 32 angular velocity bins, resulting in 256 unique trajectory candidates evaluated per planning iteration \citep{Nguyen2024Uncertainty-awareNetworks}. These primitives are precomputed in the body frame and transformed into the world frame using the fused pose estimate from VINS-Fusion. To encourage conservative planning behavior, the network is trained with an expanded safety margin around obstacles, penalizing near-collision trajectories more strongly. This adjustment biases the policy toward paths that maintain greater clearance, improving navigation robustness in dense and cluttered vegetation. In situations where only narrow clearance paths are available, the drone may enter a dead-end condition, causing it to stop and rotate in place while searching for a safe and collision-free direction to continue.\\

The CPN architecture includes an LSTM (Long Short-Term Memory) layer that encodes temporal dependencies within each trajectory, reflecting the cumulative risk associated with long-term execution. This sequence-level embedding is concatenated with the latent perception vector and passed through a series of fully connected layers, culminating in sigmoid-activated outputs that yield the probability of collision for each trajectory.\\

Ensemble learning is utilized to mitigate overfitting and improve predictive robustness. Three instances of the CPN are independently trained with different random seeds, initialization schemes, and training batch shuffles. During inference, their predicted collision probabilities are averaged to produce a consensus output, empirically reducing the variance observed in individual predictors and enhancing stability in cluttered or visually ambiguous environments. Furthermore, propagation of uncertainty in both state estimation and perception is handled through the Unscented Transform (UT), a deterministic sampling technique from the family of sigma-point filters. UT is applied to the input state vector to generate a set of representative samples capturing the nonlinear effects of uncertainty on the predicted outcomes. By evaluating the CPN across these sigma points and aggregating the results, the system provides a more accurate approximation of expected collision risk under measurement and model uncertainty, a critical feature for under-canopy flight where GPS signals are unavailable and depth data may be noisy or incomplete.\\

The decision-making module operates at 30 Hz, synchronized with depth acquisition and latent encoding. This real-time operation enables dynamic replanning in response to fast-approaching obstacles and transient environmental changes, such as moving foliage or sudden exposure gaps. Real-world tests indicate that this responsiveness is critical for avoiding collisions with thin branches and partially occluded tree trunks, particularly under low-contrast lighting conditions prevalent in forest environments.

\subsubsection{Geometric Safety Supervisor}

To ensure safe navigation, an algorithm that behaves as a supervisor evaluates the feasibility of each candidate velocity command by checking for obstacles in the depth image. This additional supervisory layer is necessary because relying solely on the learned Collision Prediction Network does not always guarantee collision-free trajectories. Although the CPN is trained to avoid obstacles, in practice was observed how it occasionally proposes unsafe velocity commands, especially in previously unseen or highly cluttered environments. An independent geometric supervisor acts as a fail-safe mechanism, filtering out unsafe commands that the CPN might generate and providing a formal guarantee of safety.


The depth image is divided into a grid of \textit{yaw} and \textit{vertical} sectors. Each sector $(i,j)$ is marked as \textit{blocked} if a significant portion of pixels correspond to obstacles within a minimum safety distance $d_\text{min}$:

\begin{equation}
B_{i,j} =
\begin{cases}
1, & \text{if } \frac{|\{(u,v) \in \text{sector}_{i,j} \mid d(u,v) < d_\text{min}\}|}{|\text{sector}_{i,j}|} > \epsilon \\
0, & \text{otherwise}
\end{cases}
\end{equation}

where $B_{i,j}$ is the blocked indicator, $d(u,v)$ is the depth value at pixel $(u,v)$, and $\epsilon$ is a threshold ratio of obstacle pixels.  

To account for the physical size of the drone and avoid collisions with obstacles, the system projects the drone into the space of each detected obstacle. The radius $r$ of the drone, which includes a safety margin, and the obstacle distance $d$ are used to compute the angular span around the obstacle where a collision would occur. This angular span is converted into a number of neighboring sectors in yaw and vertical directions that must be marked as blocked.

Formally, the number of sectors to block is computed as:

\begin{equation}
m_{i,j}^{\text{yaw}} = \left\lceil \frac{ \arctan(r/d) }{\delta_\text{yaw}} \right\rceil, \quad
m_{i,j}^{z} = \left\lceil \frac{h_\text{margin}/2}{\delta_z} \right\rceil
\end{equation}

where $\delta_\text{yaw}$ and $\delta_z$ are the angular sizes of each yaw and vertical sector, respectively, and $h_\text{margin}$ represents the vertical size of the drone. The final blocked mask including margins is obtained by expanding each detected obstacle sector across its horizontal and vertical neighbors:

\begin{equation}
\tilde{B}_{i,j} = \bigvee_{k=-m_{i,j}^{\text{yaw}}}^{m_{i,j}^{\text{yaw}}} \bigvee_{l=-m_{i,j}^{z}}^{m_{i,j}^{z}} B_{i+k,j+l}
\end{equation}

This procedure ensures that all directions leading to potential collisions, even if only partially intersecting the drone, are safely filtered out.

Finally, only velocity commands corresponding to sectors not marked as blocked are considered safe:

\begin{equation}
\mathcal{A}_\text{safe} = \{ (v_y, v_z) \mid \tilde{B}_{i(v_y), j(v_z)} = 0 \}
\end{equation}

\begin{figure}[ht!]
    \centering

    \begin{subfigure}{0.48\columnwidth}
        \centering
        \includegraphics[width=\linewidth]{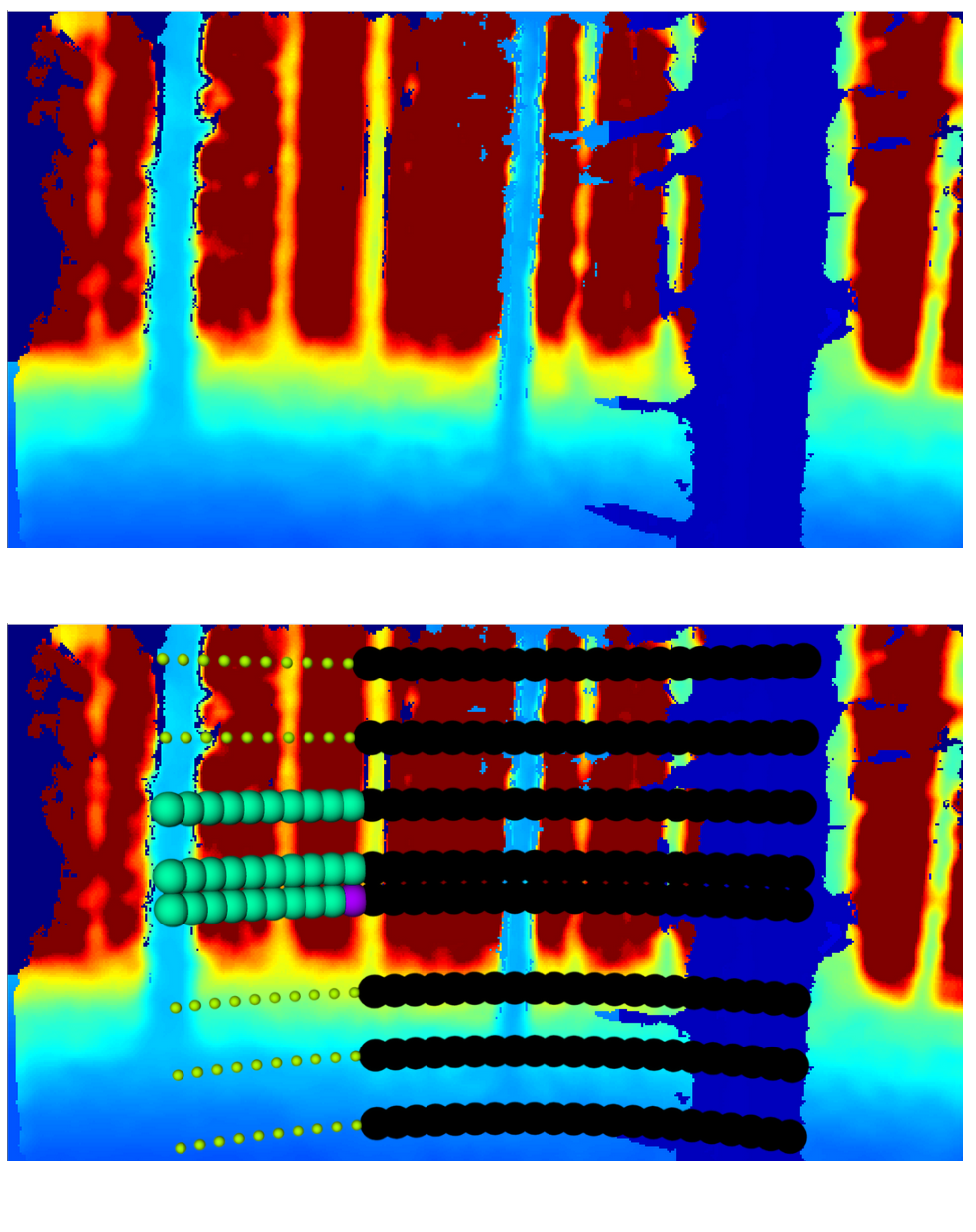}
        \caption{}
        \label{fig:planner_10}
    \end{subfigure}
    \hfill
    \begin{subfigure}{0.48\columnwidth}
        \centering
        \includegraphics[width=\linewidth]{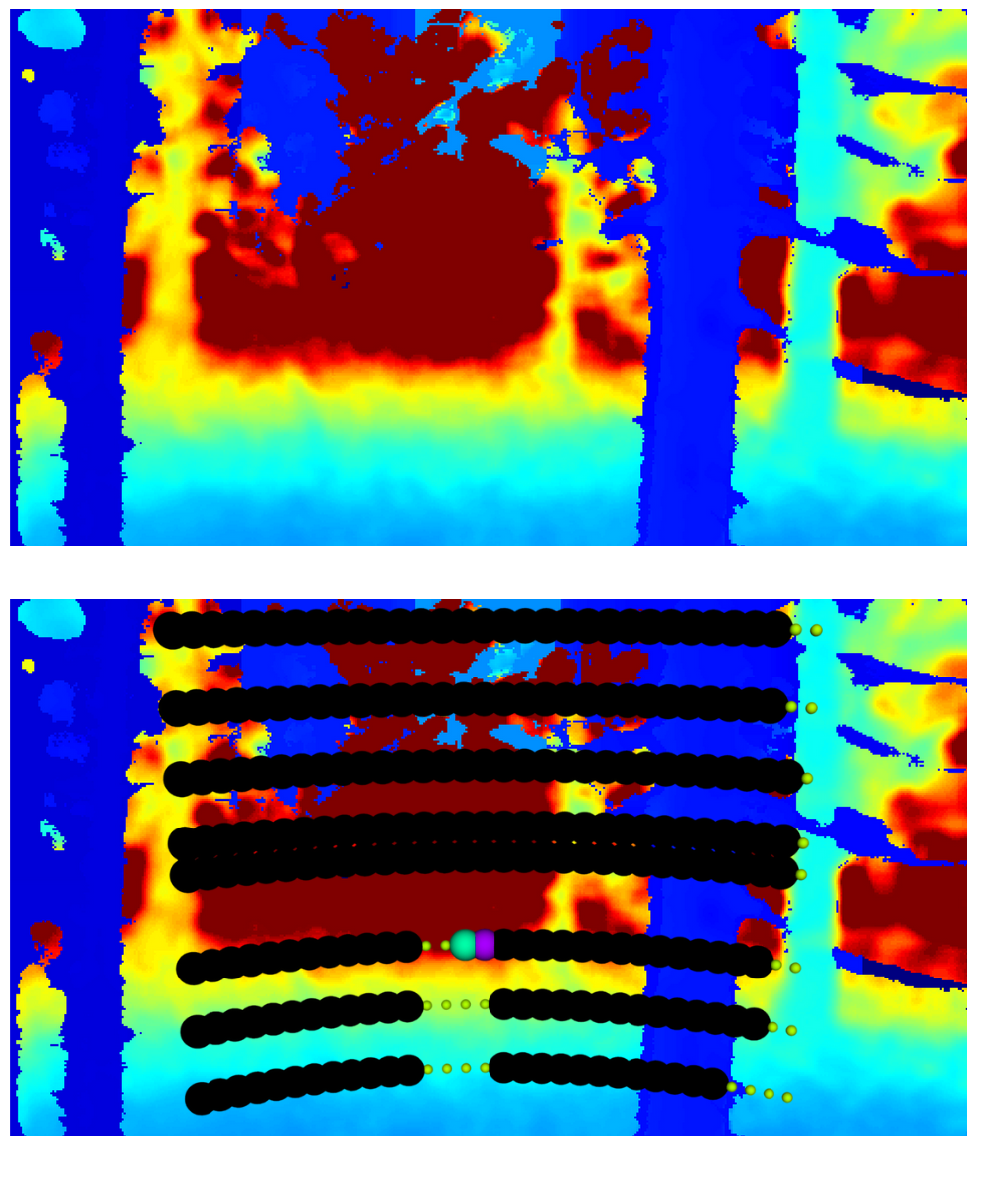}
        \caption{}
        \label{fig:planner_20}
    \end{subfigure}

    \caption{Planner real-time representation examples. Green indicates safe actions, violet shows the selected action, and black marks supervisor-blocked actions. Panels (a) and (b) illustrate different scenarios, highlighting the role of the supervisor in removing unsafe actions. In each example the top image shows the raw depth and the bottom image visualizes the decisions from the planner. The planner selects the violet action from the green safe set as the one most closely aligned with the goal direction, while the supervisor blocks all potentially colliding actions with a safety margin.
 }
    \label{fig:drone_planner}
\end{figure}

This supervisor mechanism guarantees that the drone avoids collisions while allowing the largest feasible set of velocity commands toward the goal.
Figure \ref{fig:drone_planner} provides an illustrative example of how the supervisor operates in practice.

\subsubsection{State Estimation and Sensor Fusion} 

Robust state estimation is essential for real-time control in forested environments, where visual degradation, motion blur, and low-texture scenes frequently compromise conventional odometry. In the original work by \citet{Nguyen2024Uncertainty-awareNetworks}, drone state estimation relied on the Intel RealSense T265 tracking camera. However, extensive preliminary testing revealed that this solution suffered from significant drift accumulation under dense canopy conditions, leading to unreliable pose estimates during extended flights. As a result, this approach was deemed unsuitable for the experimental scenarios considered in this study.

To overcome these limitations, the system adopts VINS-Fusion \citep{Qin2018VINS-Mono:Estimator}, an optimization-based visual-inertial odometry framework that operates directly on stereo imagery and inertial measurements from the PX4 autopilot. VINS-Fusion provides tightly coupled state estimation by jointly optimizing visual feature observations and inertial data, resulting in improved robustness against drift in environments characterized by limited visual structure and intermittent depth degradation. VINS-Fusion was configured to use the same sensor employed for depth estimation, namely the Intel RealSense D435i. Specifically, the infrared stereo image stream from the camera was used as visual input for the visual–inertial odometry pipeline.

Accurate calibration of the camera–IMU system was critical to achieving reliable performance. An initial estimate of intrinsic and extrinsic parameters was obtained using the Kalibr toolbox, providing a consistent baseline for sensor alignment \citep{furgale2013Unified}. These parameters were subsequently refined using the automatic extrinsic calibration refinement procedure available within VINS-Fusion. This process required the acquisition of extensive calibration datasets involving diverse motion patterns to sufficiently excite all degrees of freedom, ultimately leading to near-optimal alignment between the visual and inertial sensors. This calibration strategy proved essential for ensuring stable and accurate state estimation during aggressive maneuvers and prolonged autonomous flights beneath the forest canopy.

\subsubsection{Hardware and Implementation}

For the real-world implementation, a compact quadrotor was developed based on prior team designs, particularly the Learning-based Micro Flyer (LMF) framework, from previous work of \citet{Nguyen2024Uncertainty-awareNetworks}. The platform was optimized for forest navigation, featuring a 0.4 m diameter, a weight of 0.96 kg, and improved sensor integration. Depth informations are captured using a Realsense D435i camera at 480$\times$270 resolution and 30 fps, balancing perceptual accuracy and computational efficiency. The same camera images were also used as an input for the drone state estimation. The final velocity command from the selected motion primitive is transmitted to the PX4-compatible Pixhawk 6C Mini flight controller (Holybro, Hong Kong, China) via MAVROS, where the low-level controller manages attitude stabilization and thrust regulation. All computational modules run on a Jetson Orin NX onboard computer (Nvidia, Santa Clara, California, USA), which supports TensorRT-optimized inference for both the autoencoder \citep{Kulkarni2023Semantically-enhancedRobots} and the CPN \citep{Nguyen2024Uncertainty-awareNetworks}. TensorRT acceleration reduces latency by over 30\%, enabling full perception–planning–control cycles at rates exceeding 10~Hz, which is critical for high-speed maneuvering through dense vegetation without sacrificing prediction accuracy. The hardware selection prioritizes lightweight, compact design and computational efficiency to enable safe, agile navigation in dense forest environments.

\begin{figure*}[t]
  \centering
  \includegraphics[width=0.85\textwidth]{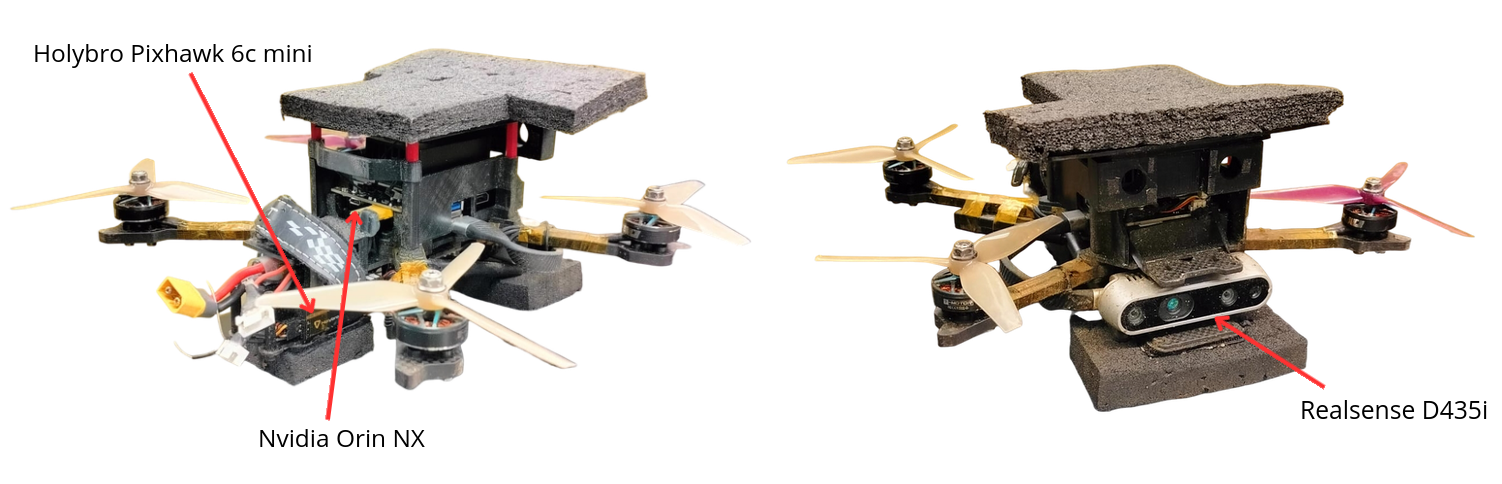}
  \vspace{2mm}
  \includegraphics[width=0.85\textwidth]{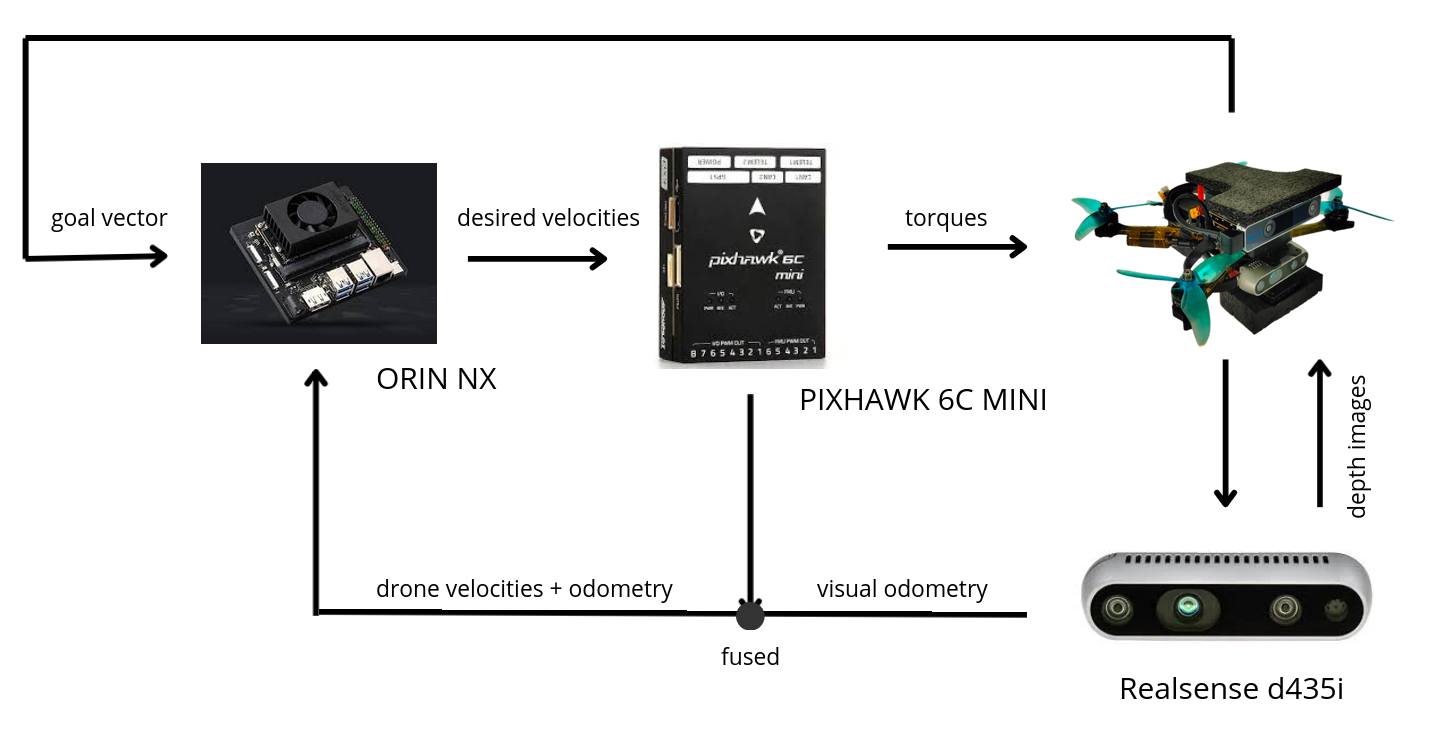}
  \caption{Quadrotor Hardware Components.}
  \label{fig:hw}
\end{figure*}

A schematic of the drone hardware is provided in Figure. \ref{fig:hw}.

\subsubsection{Implementation Details and Data Management}


Once the hardware platform was finalized, data were collected to train the Collision Prediction Network (CPN) and the autoencoder. Depth images, drone velocities and binary collision labels were generated primarily using the RotorS simulator, configured with randomized forest environments containing trees, branches, rocks, and man-made obstacles. The environment was regenerated after each collision, producing diverse scene geometries across hundreds of episodes. Depth maps and segmentation labels for additional autoencoder training were also obtained using the Isaac Gym simulator, which enabled large-scale parallel data generation with randomized small-vegetation models.

The collected datasets were stored in TFRecord format and processed through a unified TensorFlow pipeline that parsed, shuffled, and batched the depth maps, states, and labels. Both networks were trained using this pipeline, with the autoencoder trained for 40 epochs and the CPN for 500 epochs. All models were trained using fixed hyperparameters taken from the original work \citep{Nguyen2024Uncertainty-awareNetworks, Kulkarni2023Semantically-enhancedRobots}, with only the autoencoder batch size reduced for computational efficiency.
Specifically, the Collision Prediction Network (CPN) was trained using a learning rate of $5\times10^{-5}$ and a batch size of 64. The autoencoder was trained with a learning rate of $1\times10^{-4}$ and a batch size of 32.
Training was performed on a Dell laptop equipped with an NVIDIA RTX A4500 GPU.

\subsection{Experimental Setup and Test Environments}

To evaluate navigation performance in real-world scenarios, multiple autonomous flight campaigns were conducted in boreal forest environments without GNSS support. The drone operated fully autonomously from takeoff to landing, relying exclusively on onboard sensing, visual–inertial state estimation, and learning-based motion planning. All experiments were performed with onboard computation and sensing to reflect realistic deployment conditions.

Each experimental campaign followed a common protocol, including repeated autonomous flights along predefined forest corridors, systematic logging of depth images, odometry, and control inputs, and execution at different target forward velocities. This standardized setup provides a consistent basis for quantitative and qualitative comparison across environments and algorithms.

Three real-world forest environments with increasing structural and perceptual complexity were selected for evaluation. The Difficult Forest served as the primary benchmark due to the availability of comparable results from prior work. The Medium Forest and Very Difficult Forest were included to assess generalization in both less cluttered and extremely challenging vegetation conditions, respectively. All test sites were located in Palohein\"{a}, Helsinki, Finland (60°15'30" N, 24°55'20" E).

Forest scenes were classified according to the density-based criteria proposed by \citet{Liang2019ForestMeasurements}, which group boreal forest environments into three levels based on tree density and understory complexity. In this framework, easy forests are characterized by fewer than 700 trees/ha and minimal understory, medium forests by approximately 1000 trees/ha with sparse understory vegetation, and difficult forests by roughly 2000 trees/ha with dense understory growth. Following this categorization, the medium, difficult, and very difficult environments used in our experiments exhibited tree densities of approximately 1040 trees/ha, 2220 trees/ha and 2000 trees/ha, respectively. Tree density alone does not fully capture the complexity of the environment. Notably, the most challenging forest is not associated with the highest tree density. Despite having fewer trees, this environment contains longer, thinner, and more irregular branches, as well as slender trunks, which significantly reduce the available free space for drone navigation. Moreover, these structural characteristics accentuate perception degradation, making navigation considerably more difficult than in denser but more structured forests.

\renewcommand{\arraystretch}{4.5}

\begin{table}[ht!]
    \centering
    \renewcommand{\arraystretch}{1.5}
    \resizebox{\columnwidth}{!}{
    \begin{tabular}{l p{1.5cm} p{1.5cm} p{3.2cm} p{3.2cm}}
        \toprule
        \textbf{Env. ID} & \textbf{Type} & \textbf{Density} & \textbf{Description} & \textbf{Purpose} \\
        \midrule
        RW-1 & Difficult Forest (baseline) & 2220 trees/ha &
        Dense spruce stand with frequent low branches and limited illumination. &
        Primary benchmark enabling direct comparison with existing systems. \\
        \hline
        RW-2 & Medium Forest & 1040 trees/ha &
        More open forest with wider traversable paths. &
        Evaluation of system reliability in visually less complex environments. \\
        \hline
        RW-3 & Very Difficult Forest & 2000 trees/ha &
        Highly cluttered mixed vegetation with thin branches often below stereo depth detection, resulting in reduced navigable free space despite fewer trees than the baseline. &
        Assessment of robustness under severe perception degradation. \\
        \bottomrule
    \end{tabular}
    }
    \caption{Descriptions of the test environments and their respective evaluation purposes.}
    \label{tab:environment_description}
\end{table}

\begin{figure*}[ht!]
  \centering
  \begin{subfigure}[b]{0.32\textwidth}
      \includegraphics[width=\textwidth]{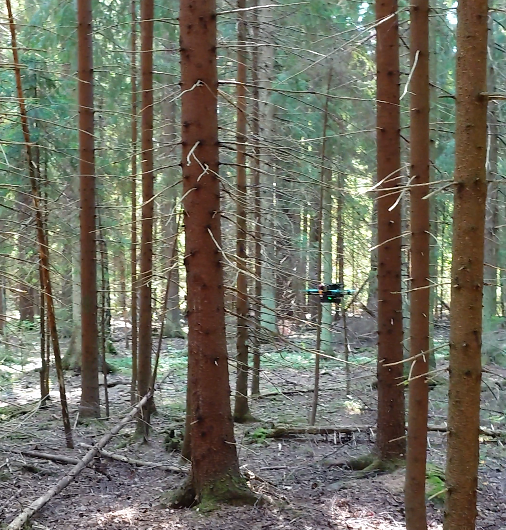}
      \caption{RW-1 (baseline)}
      \label{fig:rw1}
  \end{subfigure}
  \begin{subfigure}[b]{0.32\textwidth}
      \includegraphics[width=\textwidth]{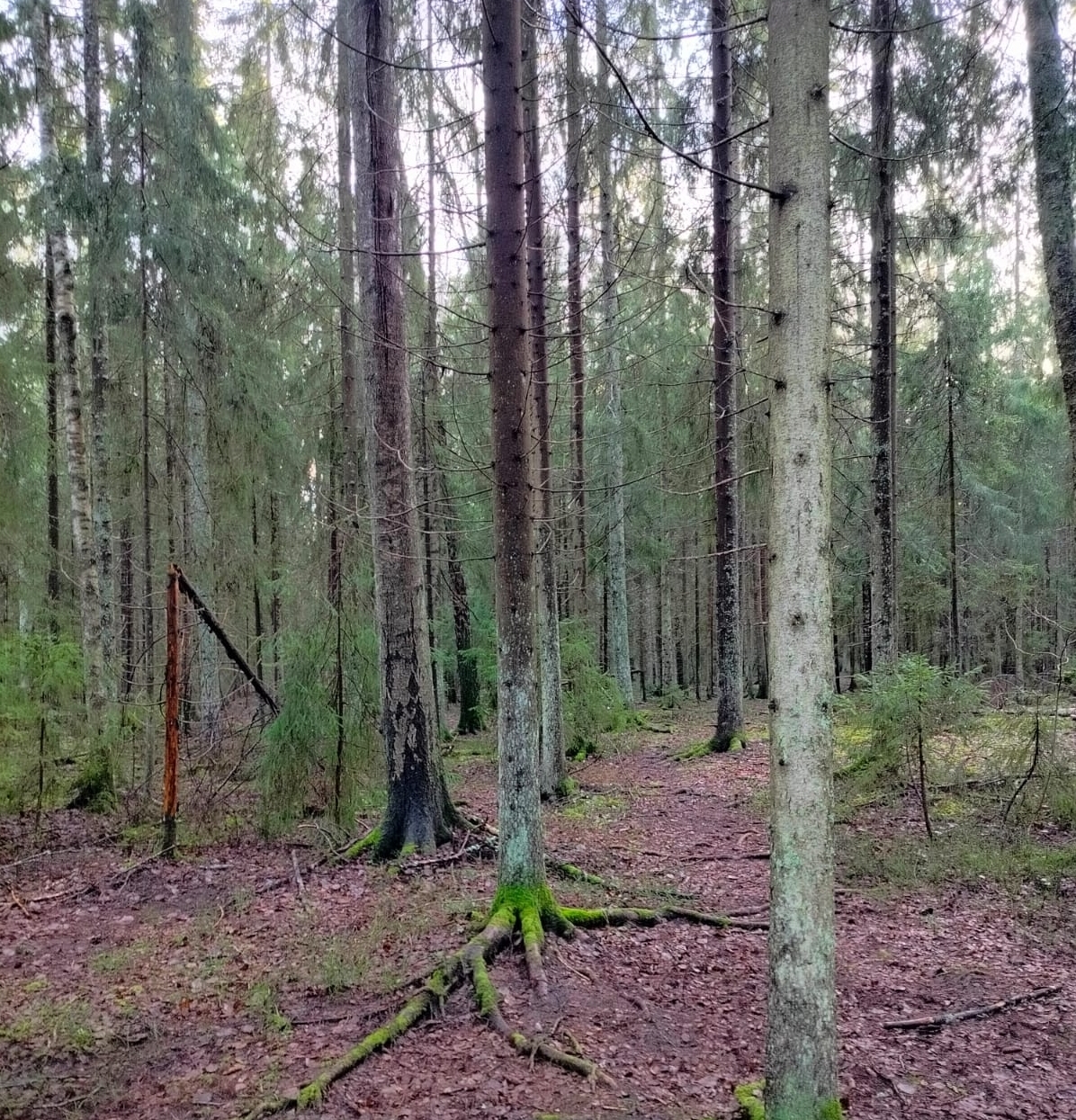}
      \caption{RW-2 (medium)}
      \label{fig:rw2}
  \end{subfigure}
  \begin{subfigure}[b]{0.32\textwidth}
      \includegraphics[width=\textwidth]{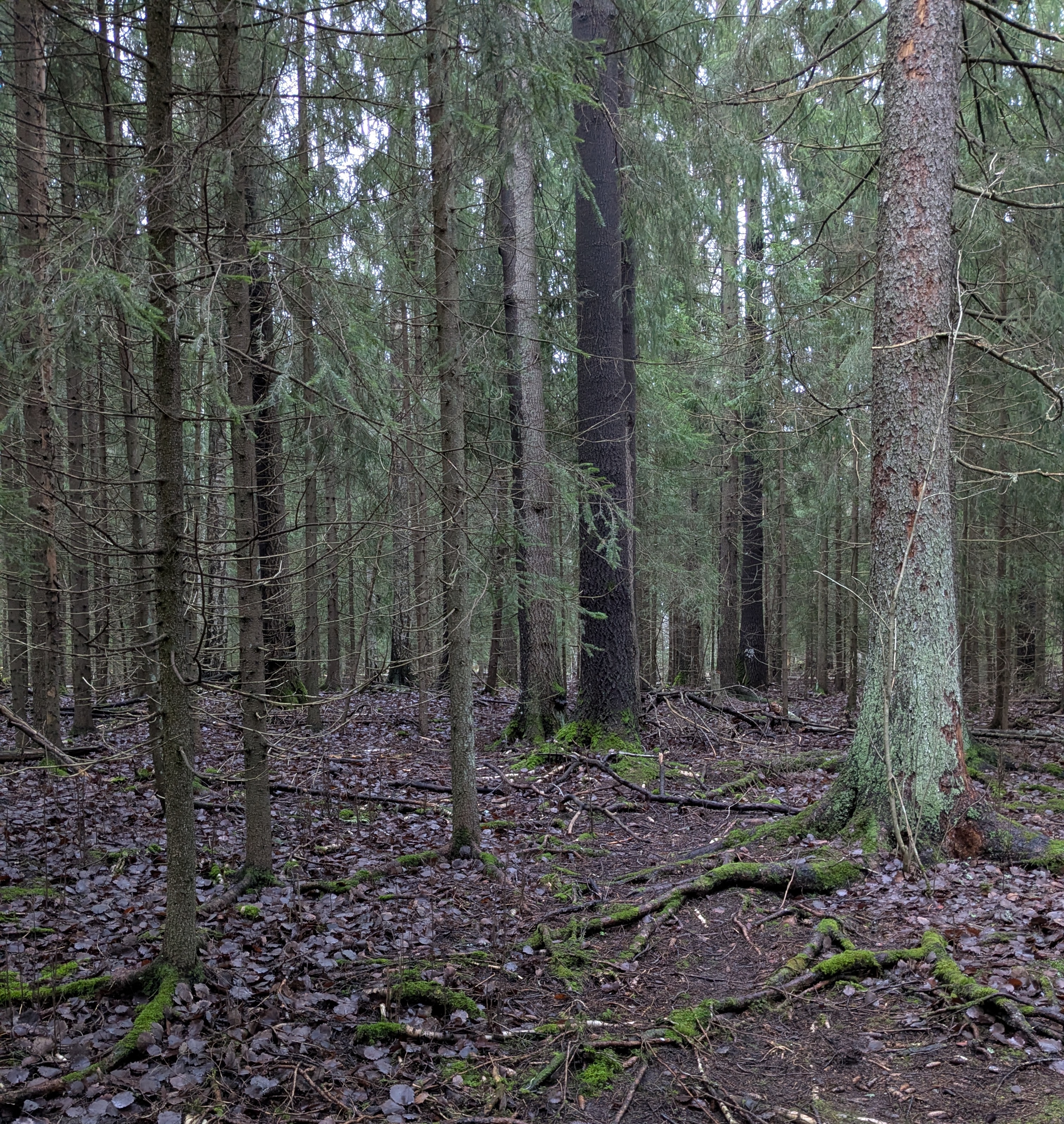}
      \caption{RW-3 (very difficult)}
      \label{fig:rw3}
  \end{subfigure}

  \caption{Examples of vegetation in the three real-world environments.}
  \label{fig:palo_all}
\end{figure*}

Representative images of the baseline, medium, and very difficult forest environments are shown in Figure~\ref{fig:palo_all}, while a detailed characterization of the three test environments is provided in Table~\ref{tab:environment_description}.

\subsubsection{Baseline: Difficult Forest}

The baseline evaluation environment consists of a dense spruce-dominated forest previously used in multiple autonomous navigation studies \citep{Karjalainen2023ARESULTS, karhunen2025fieldevaluationoptimizationlightweight, DelColGuglielmo2024Autonomous}. Strong canopy closure substantially reduces illumination at flight altitude, while abundant low branches and undergrowth create narrow, visually cluttered flight corridors. The estimated vegetation density is approximately 2220 trees/ha, resulting in frequent occlusions and limited safe passage widths. These characteristics make the site a representative and challenging benchmark for under-canopy autonomous flight.

\subsubsection{Medium Forest}

The medium-complexity environment is characterized by reduced vegetation density and wider traversable corridors compared to the baseline forest. Obstacles are more sparsely distributed, and visual conditions are generally more favorable. This environment is used to verify that the navigation system maintains stable behavior, accurate state estimation, and smooth trajectory execution under less demanding perceptual conditions, without introducing planner instability or control oscillations.

\subsubsection{Very Difficult Forest}

The very difficult forest represents the most challenging test condition. The area contains high-density mixed vegetation including spruce, pine, bushes, and thin, irregular branches occupying a large portion of the navigable space. Despite a slightly lower nominal tree density than the baseline forest, thicker trunks, irregular branch geometry, and near-field foliage significantly increase perceptual complexity. Many obstacles fall below the reliable sensing threshold of the stereo depth camera, leading to frequent depth noise, occlusions, and partial perception failures. This environment is used to assess the practical operational limits of the perception and planning pipeline under severe sensor degradation.

\subsubsection{Algorithms Tested and Compared}

Four navigation configurations were evaluated across the defined environments using a common experimental protocol. All experiments consisted of fully autonomous 60\,m flights, repeated 15 times per configuration, with identical start and goal positions, sensing hardware, and control interfaces to ensure a fair comparison.

Two baseline configurations were based on the SEVAE-Oracle framework \citep{Nguyen2024Uncertainty-awareNetworks}. The first employed the original SEVAE-Oracle architecture without modification, providing a historical reference under identical environmental conditions. The second baseline used a fine-tuned version of SEVAE-Oracle \citep{DelColGuglielmo2024Autonomous}, retrained with data collected in the baseline forest and representing the strongest previously reported system at this site. Both baseline configurations were evaluated exclusively in the Difficult Forest environment to enable direct and controlled comparison with the proposed method. Both the baseline systems were tested at the nominal speed of 1.0\,m/s

The proposed enhanced navigation system, referred to as DeFoP (Deep Forest Pilot), was evaluated in all three environments. In the Difficult Forest (baseline), DeFoP was tested at two target forward velocities: 1.0\,m/s, representing a practical operating speed for forest navigation, and 1.3\,m/s, designed to probe robustness under reduced reaction time and increased motion constraints. In the Medium Forest and Very Difficult Forest, DeFoP was evaluated at 1.0\,m/s to assess generalization and robustness under differing perceptual challenges.

In addition to the internally evaluated baselines, the results of all tested configurations were compared against previously published state-of-the-art methods by \citet{Karjalainen2023ARESULTS} and \citet{karhunen2025fieldevaluationoptimizationlightweight} using performance metrics reported in their respective studies. These comparisons were conducted in the Difficult Forest environment (baseline) for both external methods, while an additional comparison with the approach of \citet{karhunen2025fieldevaluationoptimizationlightweight} was included for the Medium Forest, where comparable experimental results were available.

By executing all configurations under identical environmental and operational conditions, observed performance differences can be attributed to algorithmic design choices rather than external factors.

\subsection{Performance Assessment}

System performance is evaluated along four complementary dimensions that collectively characterize navigation reliability, motion quality, robustness in cluttered environments, and real-time feasibility. Navigation reliability is assessed by measuring the ability of the drone to complete predefined forest corridors without collisions or activation of emergency procedures. Each trial is considered successful when the vehicle reaches the final waypoint from the starting location while maintaining a safe trajectory, providing a direct indicator of the stability and consistency of the perception and planning pipeline under representative operating conditions.

Trajectory and motion quality are evaluated using kinematic quantities derived from onboard odometry, including average forward velocity, the discrepancy between commanded and executed motion, linear acceleration profiles, and the smoothed root mean square acceleration. These metrics enable identification of oscillatory behavior, inefficient motion patterns, and abrupt control actions. Particular attention is given to the effects of the planning stabilization mechanism and the inclusion of a lateral velocity component in the motion primitives, both of which are designed to improve responsiveness and directional consistency during flight in dense clutter.

Robustness in complex forest environments is examined through qualitative and quantitative observations in scenes containing thin structures, partially occluded branches, and noisy or incomplete depth measurements. The evaluation considers the continuity of obstacle representations in the processed depth maps, the temporal consistency of the latent features produced by the autoencoder, the clearance maintained around obstacles along executed trajectories, and the frequency of situations in which the drone approaches obstacles more closely than desired. These observations provide insight into how effectively the perception pipeline preserves fine geometric detail and how the collision prediction module promotes conservative and safe navigation behavior.

Finally, real-time system performance is assessed through timing statistics collected for the main components of the pipeline. This includes the execution time of depth preprocessing, autoencoder inference, collision prediction, and the geometric safety supervisor, as well as the overall perception–planning–control cycle frequency. Computational resource utilization on the Jetson Orin NX is monitored to verify that the system satisfies the temporal constraints required for agile autonomous navigation in dense forest environments.

\section{Results}

\subsection{Baseline (Difficult Forest) Results}

In Table \ref{tab:combined_navigation} a comparison is presented between all the systems described in this section, tested in the same environment.

\begin{table*}[ht!]
\centering

\caption{Navigation performance comparison over 15 flights in different forest difficulties: (a) baseline difficult forest, (b) medium forest, (c) very difficult forest.}
\resizebox{\textwidth}{!}{
\begin{tabular}{lccccccc}
\hline
\textbf{\huge Method} & 
\textbf{\huge Target Vel. (m/s)} & 
\textbf{\huge Dist. (m)} &
\textbf{\huge Avg. Disp. Vel. (m/s)} &
\textbf{\huge Avg. Odom. Vel. (m/s)} &
\textbf{\huge Avg. Acc. (m/s$^2$)} &
\textbf{\huge Smooth. RMS Acc.} &
\textbf{\huge Success Rate} \\ 
\hline
\multicolumn{8}{c}{\huge \textbf{(a) Baseline Difficult Forest}} \\ \hline

\textbf{\huge DeFoP (Proposed Method)} &
\textbf{\huge 1} &
\textbf{\huge 59.0454 $\pm$ 1.2942} &
\textbf{\huge 0.8183 $\pm$ 0.0784 }&
\textbf{\huge 0.9022 $\pm$ 0.0424} &
\textbf{\huge 0.6977 $\pm$ 0.2473} &
\textbf{\huge 3.3080 $\pm$ 5.9800} &
\textbf{\huge 15/15} \\

\textbf{\huge DeFoP (Proposed Method)} &
\textbf{\huge 1.3} &
\textbf{\huge 58.8642 $\pm$ 0.7398}&
\textbf{\huge 0.9851 $\pm$ 0.0842} &
\textbf{\huge 1.0951 $\pm$ 0.0768} &
\textbf{\huge 1.5798 $\pm$ 1.3147} &
\textbf{\huge 16.7440 $\pm$ 26.7915} &
\textbf{\huge 15/15} \\

\huge seVAE--ORACLE \citep{Nguyen2024Uncertainty-awareNetworks} &
\huge 1 &
\huge 41.8886 $\pm$ 16.1898 &
\huge 0.9512 $\pm$ 0.0405 &
\huge 1.0108 $\pm$ 0.0110 &
\huge 0.5842 $\pm$ 0.3989 &
\huge 4.9514 $\pm$ 7.8283 &
\textbf{\huge 5/15} \\

\huge Fine-tuned seVAE \citep{DelColGuglielmo2024Autonomous} &
\huge 1 &
\huge 44.3038 $\pm$ 17.1174 &
\huge 0.9372 $\pm$ 0.0635 &
\huge 1.0012 $\pm$ 0.0355 &
\huge 0.9290 $\pm$ 0.7590 &
\huge 10.9661 $\pm$ 13.3143 &
\textbf{\huge 8/15} \\

\huge Ego-Planner \citep{Karjalainen2023ARESULTS} &
\huge 1 &
\huge / &
\huge / &
\huge / &
\huge / &
\huge / &
\textbf{\huge 9/19} \\

\huge LTA-OM SLAM + IPC \citep{karhunen2025fieldevaluationoptimizationlightweight} &
\huge 1 &
\huge 57 &
\huge 0.7 &
\huge 0.75 &
\huge / &
\huge / &
\textbf{\huge 15/15} \\

\huge LTA-OM SLAM + IPC \citep{karhunen2025fieldevaluationoptimizationlightweight} &
\huge 2 &
\huge 57 &
\huge 1.05 &
\huge 1.3 &
\huge / &
\huge / &
\textbf{\huge 5/15} \\

\hline
\multicolumn{8}{c}{\textbf{\huge (b) Medium Forest}} \\ \hline

\textbf{\huge DeFoP (Proposed Method)} &
\textbf{\huge 1} &
\textbf{\huge 59.0454 $\pm$ 1.2942} &
\textbf{\huge 0.8183 $\pm$ 0.0784} &
\textbf{\huge 0.9022 $\pm$ 0.0424} &
\textbf{\huge 0.6977 $\pm$ 0.2473} &
\textbf{\huge 3.3080 $\pm$ 5.9800} &
\textbf{\huge 15/15} \\

\huge LTA-OM SLAM + IPC \citep{karhunen2025fieldevaluationoptimizationlightweight} &
\huge 1 &
\huge 57 &
\huge 0.76 &
\huge 0.81 &
\huge / &
\huge / &
\textbf{\huge 12/15} \\

\hline
\multicolumn{8}{c}{\textbf{\huge (c) Very Difficult Forest}} \\ \hline

\textbf{\huge DeFoP (Proposed Method)} &
\textbf{\huge 1} &
\textbf{\huge 58.5077 $\pm$ 0.3954} &
\textbf{\huge 0.9227 $\pm$ 0.0395} &
\textbf{\huge 0.9571 $\pm$ 0.0221} &
\textbf{\huge 0.5202 $\pm$ 0.3791} &
\textbf{\huge 4.3148 $\pm$ 7.7644} &
\textbf{\huge 12/15} \\

\hline
\end{tabular}
}

\label{tab:combined_navigation}
\end{table*}

\subsubsection{SEVAE-ORACLE (Original Implementation)}

The original SEVAE-ORACLE \citep{Nguyen2024Uncertainty-awareNetworks} was evaluated first in the baseline dense forest environment to establish a baseline for comparison. Of the 15 autonomous flights performed, only 5 resulted in complete mission success, corresponding to a success rate of 33\% (Table \ref{tab:combined_navigation} a). This reflects the difficulty of the operating environment, characterized by extremely low visibility, thin branches, and irregular vegetation densities.

Across all flights, the system achieved an average displacement, defined as the distance between the start and end points of each flight, of 41.89 m (±16.19), with significant variation between runs. Although navigation performance deteriorated in several flights, velocity tracking remained stable: the average velocity derived from displacement was 0.95 m/s (±0.04), closely aligned with the odometry estimate of 1.01 m/s (±0.01), as shown in Table \ref{tab:combined_navigation} a. This indicates that the low-level inner-loop control from the px4 controller, maintained speed tracking reliably even when perception-based navigation failed.

Acceleration-based smoothness indicators exhibited large variability. The average standard deviation of acceleration was 4.86 (±7.86), and Root Mean Square (RMS) acceleration showed similar dispersion (4.95 ±7.83). These values indicate that while the controller occasionally executed smooth and stable trajectories, several flights required late corrective maneuvers, resulting in abrupt accelerations and nonlinear motion responses.

Overall, the original SEVAE-ORACLE \citep{Nguyen2024Uncertainty-awareNetworks} was capable of traversing parts of the dense forest but struggled to perceive and anticipate critical forest geometry, particularly thin branches and cluttered low-visibility vegetation. The large variance in flight performance is consistent with the limited representational fidelity of the baseline model under highly unstructured sensory conditions.

In Table \ref{tab:flight_results_set5} all the results of flights from this experiment are presented.

\subsubsection{SEVAE-ORACLE Fine-Tuned on Synthetic Forest Data}

The fine-tuned version of SEVAE-ORACLE \citep{DelColGuglielmo2024Autonomous} was intended to improve scene understanding due to exposure to real-world forest imagery during additional training. While results showed partial improvement, only 8 out of 15 flights were completed successfully, corresponding to a success rate of 53\%, representing a notable but still insufficient improvement over the baseline.

The average displacement increased modestly to 44.30 m (±17.12), although large dispersion indicates persistent inconsistency between flights. Similar to the original implementation, velocity tracking remained accurate and stable: the average displacement-derived speed was 0.94 m/s (±0.06), while odometry estimated a mean of 1.00 m/s (±0.04), as shown in Table \ref{tab:combined_navigation} a.

Acceleration variability increased relative to the original model. Standard deviation of acceleration increased to 10.87 (±13.34), with RMS acceleration nearly identical at 10.97 (±13.31). This indicates that while the improved representation enabled more reliable obstacle detection, the controller often reacted late, triggering sharper and more aggressive avoidance maneuvers rather than earlier and smoother corrective action.

In effect, fine-tuning strengthened sensitivity to obstacles but did not allow early anticipation of collisions sufficient to produce stable control trajectories. The system improved detection but did not consistently translate this improvement into smoother, more confident flight behavior.

In Table \ref{tab:flight_results_set4} all the results of flights from this experiment are presented.

\subsubsection{DeFoP Method}

At a nominal speed of 1.0 m/s the system demonstrated the highest consistency and reliability across all experiments. All 15 flights were successful achieving a 100 percent success rate and significantly outperforming both SEVAE baselines under identical conditions.

Displacement remained consistently high with most flights reaching the full 60 meter test trajectory with minimal deviation. The average velocity computed from displacement ranged between 0.84 and 0.88 m/s while odometry reported values between 0.87 and 0.95 m/s indicating accurate and stable velocity tracking throughout the runs.

Control smoothness metrics also improved substantially. While occasional flights recorded moderate acceleration peaks during sudden obstacle encounters most missions exhibited low standard deviation and RMS acceleration values typically between 0.66 and 0.91 indicating controlled and stable motion. These results suggest that the enhanced semantic representation and heuristics for resolving planning indecision enabled earlier and more informed trajectory adjustments.

Overall the 1.0 m/s experiment demonstrated that the proposed method can consistently navigate extremely cluttered forests with smooth and stable control performance outperforming existing methods in both mission reliability and flight quality. In Table \ref{tab:flight_results_set1} all the results of flights from this experiment are presented and in Figure \ref{fig:all_paths_a} the paths followed by the drone during these flights are shown.

To evaluate the dynamic limits of the system the commanded forward speed was increased to 1.3 m/s. Despite reduced reaction time and increased demand on perception and planning all 15 flights were completed successfully maintaining a 100 percent success rate under more challenging dynamic conditions.

The higher speed resulted in increased acceleration variability and more reactive control behavior. Although displacements remained near the full mission length the system required sharper corrective maneuvers when encountering obstacles as the reduced temporal margin limited the ability to shape trajectories gradually. Smoothness metrics measured by RMS acceleration increased accordingly reflecting the higher physical demand of faster flight.

Performance remained superior to both SEVAE-ORACLE variants as shown in Table \ref{tab:combined_navigation} a. The results indicate that the proposed architecture generalizes effectively to higher-speed operation while maintaining full mission success despite more aggressive flight profiles.

These experiments establish an operational trade-off. While 1.3 m/s is fully achievable 1.0 m/s represents the optimal balance between safety smoothness and anticipatory control. The system remains capable at higher speeds but operates closer to the real-time limits of perception and planning. In Table \ref{tab:flight_results_set3} all the results of flights from this experiment are presented.

\subsection{Medium Forest Results}

DeFoP was tested in the Medium forest at 1.0 m/s target speed.

All 15 missions from the second environment were completed successfully, resulting in a 100 \% success rate, and no operational issues or anomalies were observed throughout the campaign. This makes the medium-density forest dataset a clean, fully successful reference case to evaluate the capabilities of the proposed navigation framework.

Across all flights, the drone consistently reached the end of the test corridor, with an average displacement of 58.51 m (± 0.40), showing that every run closely matched the full intended path without premature stopping. The average velocity computed from displacement was 0.923 m/s (± 0.040), and the odometry-derived velocity averaged 0.957 m/s (± 0.022), indicating accurate and stable velocity tracking throughout the trajectories (Table \ref{tab:combined_navigation} b).

Acceleration-based smoothness metrics further highlight the improved control characteristics observed in this environment. The mean acceleration magnitude was 0.520 $\text{m/{s}}^2$ (± 0.379), while the standard deviation of acceleration averaged 4.23 (± 7.79), with RMS acceleration exhibiting similar values at 4.31 (± 7.76). These metrics are substantially lower than those recorded in the difficult and very difficult forest, reflecting the benefits of reduced clutter and more predictable free-space geometry, which allowed the planner to act more anticipatorily and less reactively (Table \ref{tab:combined_navigation} b). This is further supported by the flight trajectories in this environment, which are more consistent across runs and follow straighter paths toward the goal, as illustrated in Figure \ref{fig:all_paths_b}.

Only a few individual flights (notably Flights 3, 6, 10, and 15) showed localized peaks in acceleration due to brief reactive maneuvers when sudden narrow obstacles entered the camera field of view, as shown in Table \ref{tab:flight_results_set6}. Even in these situations, the system rapidly stabilized and resumed smooth forward motion, demonstrating robust recovery dynamics.




\subsection{Very Difficult Forest Results}



DeFoP was tested in the Very Difficult forest at 1.0 m/s target speed.
Due to higher complexity and limited visibility of thin obstacles, navigation proved considerably more difficult. Twelve flights out of fifteen successfully reached the goal, corresponding to an 80 \% completion rate, while three runs resulted in premature termination due to collision or inability to proceed safely \ref{tab:flight_results_set2}.

Despite the higher environmental complexity, the successful flights demonstrated stable motion behavior, with an average displacement of 52.3 m (± 14.3) and a mean velocity along the displacement vector of 0.78 m/s (± 0.10). An average speed from odometry measurements was 0.90 m/s (± 0.05), indicating that despite increased evasive maneuvers and replanning, forward motion remained close to the nominal speed. Acceleration metrics showed higher variability compared to the baseline environment, with an average of 1.07 $\text{m/{s}}^2$ (± 0.68), consistent with the sharper directional changes required to avoid branches and dense vegetation. Smoothness measures confirmed this trend, with both standard deviation and RMS values reflecting the more abrupt maneuvering inherent to the terrain. Collectively, these results show that the system was able to navigate reliably in an environment notably harsher than the baseline, maintaining a solid success rate despite pronounced clutter and reduced free-space predictability.

\begin{figure*}[ht!]
    \centering
    \begin{subfigure}[b]{\textwidth}
        \includegraphics[width=\textwidth]{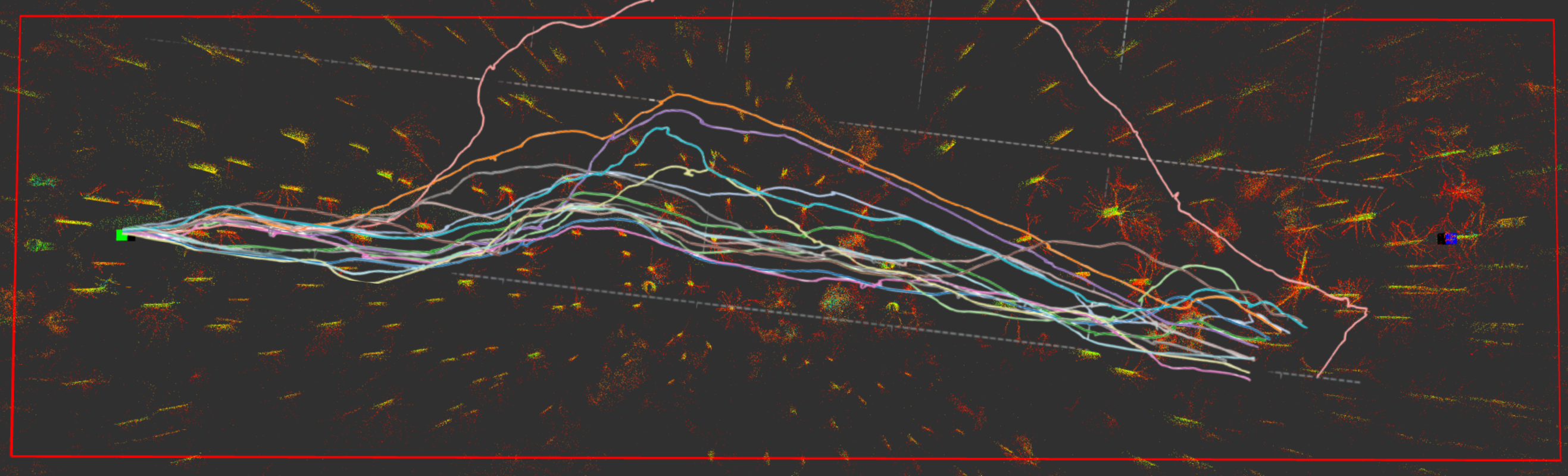}
        \caption{}
        \label{fig:all_paths_a}
    \end{subfigure}
    
    \begin{subfigure}[b]{\textwidth}
        \includegraphics[width=\textwidth]{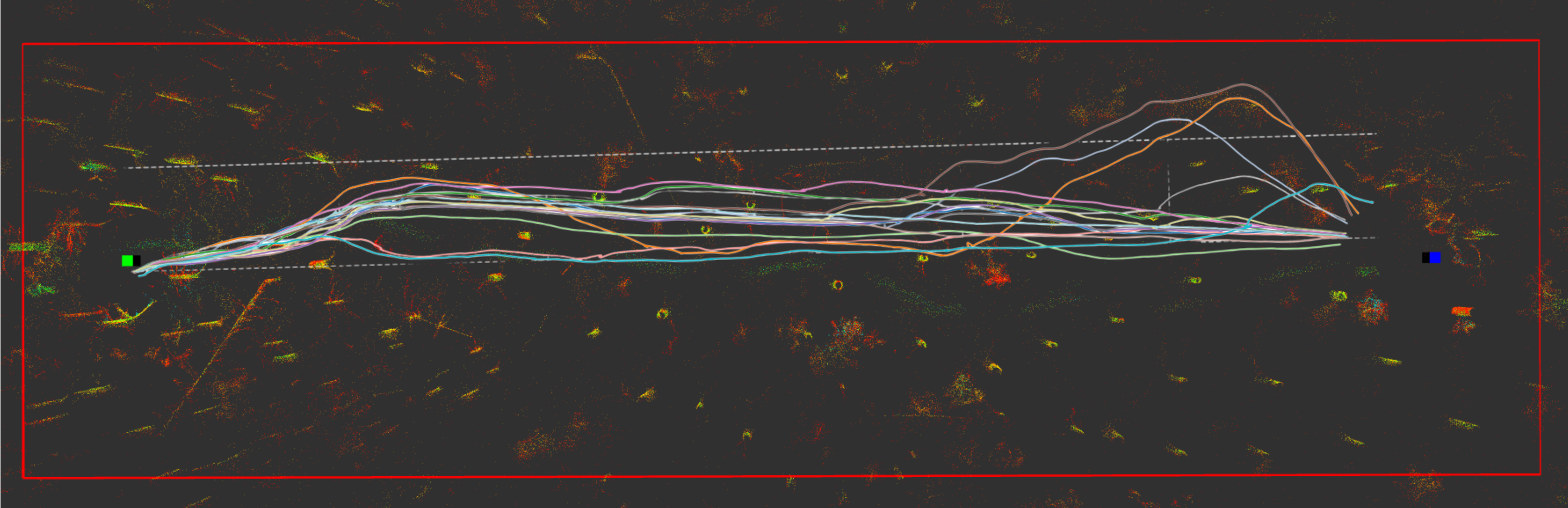}
        \caption{}
        \label{fig:all_paths_b}
    \end{subfigure}
    
    \begin{subfigure}[b]{\textwidth}
        \includegraphics[width=\textwidth]{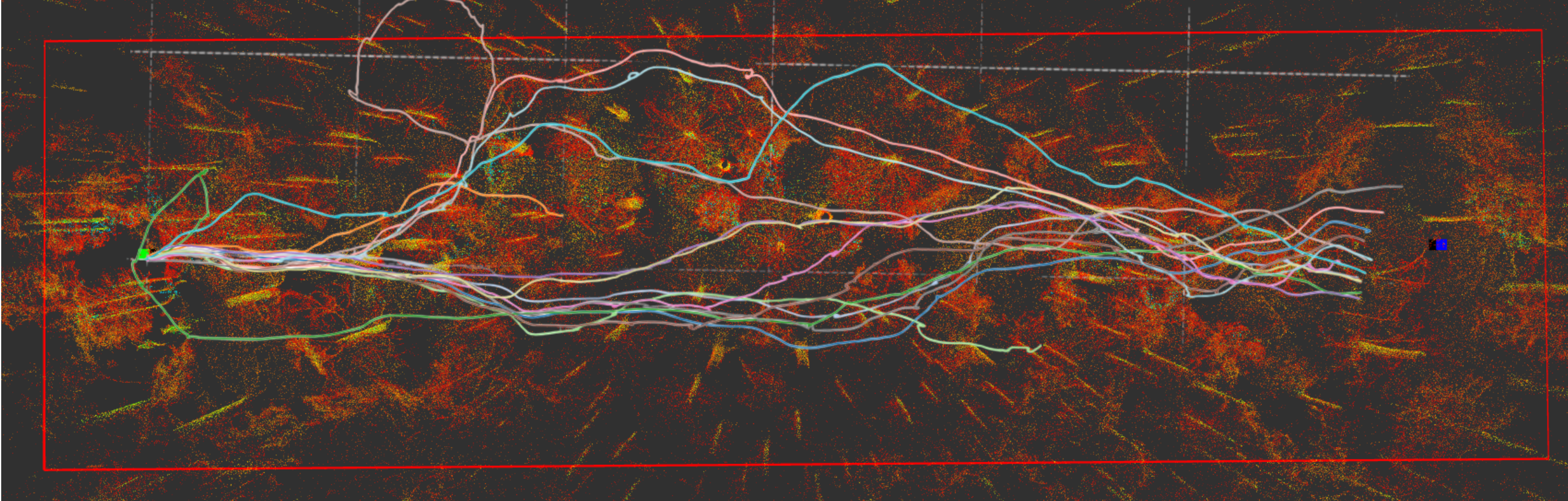}
        \caption{}
        \label{fig:all_paths_c}
    \end{subfigure}

    \caption{All flight paths using the proposed method across three forest environments at 1 m/s based on Vins-Fusion visual odometry. (a) Baseline difficult forest, (b) medium forest, (c) very difficult forest.}
    \label{fig:all_paths}
\end{figure*}

It is also evident in Fig. \ref{fig:all_paths_c} that, in some successful flights, the drone performed loop-like maneuvers to navigate around obstacles after entering local dead-end regions. These behaviors indicate that the proposed system can reliably recover from challenging situations and continue toward the final waypoint, demonstrating strong robustness in highly constrained forest environments.

In Table \ref{tab:combined_navigation} c, a summary of the average performances of the proposed method in this last environment is presented, while detailed data regarding each flight in this environment is available in Table \ref{tab:flight_results_set2}.




\section{Discussion}

\subsection{Performance comparison}

Experimental results demonstrate that DeFoP achieves significantly greater robustness and reliability in cluttered, GNSS-denied forest environments compared to other state of the art algorithm. 

The main contribution of this work lies in bridging the gap between simulation-trained visual navigation and real-world deployment through targeted architectural and behavioral enhancements. The introduction of a supervisor module, responsible for filtering unsafe motion primitives based on real-time depth analysis, proved essential for ensuring flight safety. While the neural network–based planner predicts collision probabilities, its understanding of environmental risk can remain unreliable in ambiguous visual conditions. The supervisor compensates for this limitation by validating the planner outputs against actual sensed geometry, rejecting actions that intersect with nearby obstacles. This safeguard allows the drone to maintain stable and collision-free behavior even in visually degraded or overgrown areas.

The depth improver module further enhances perception robustness by mitigating missing or noisy depth regions, particularly around thin branches and small foreground obstacles. By applying localized kernel-based interpolation to pixels within a close-range threshold, the module reconstructs partially undefined depth regions, improving the visibility of near-field obstacles. However, corrective ability remains constrained by the underlying stereo sensor limitations: when fine structures such as small branches are entirely absent from the original depth map, the improver cannot infer their presence. Consequently, although the enhancement reduces sensor noise and partial voids, it cannot compensate for the fundamental inability of stereo depth estimation to represent subpixel-thin obstacles under challenging lighting or texture conditions.

Optimization of the CPN using TensorRT yielded a major practical improvement by increasing inference frequency from approximately 4 Hz to 10 Hz. This gain enables smoother trajectory evaluation and faster replanning, resulting in more responsive and natural obstacle avoidance behavior at higher speeds. Together with the semantically fine-tuned autoencoder, these optimizations bring the perception–planning loop closer to real-time operation under field conditions.

A detailed comparison of the methods (Table \ref{tab:combined_navigation}) across forest densities highlights the operational improvements achieved with the proposed approach. In medium and high density forest, the proposed method achieved a 100 \% success rate across all flights, demonstrating complete trajectory completion with minimal deviation, smooth acceleration profiles, and stable velocity tracking. In contrast, both SEVAE-ORACLE variants struggled under the same conditions: the original implementation \citet{Nguyen2024Uncertainty-awareNetworks} completed 5 out of 15 flights (33\% success rate), while the fine-tuned version \citet{DelColGuglielmo2024Autonomous} reached 8 out of 15 flights (53\% success rate).

Among the systems tested and compared in this journal, the only approach with comparable results in the baseline forest is the method of \citet{karhunen2025fieldevaluationoptimizationlightweight}, which achieved 15 out of 15 successful flights; however, this performance was obtained at a substantially lower commanded velocity of 0.75 m/s (Table \ref{tab:combined_navigation}). When operated at a higher real velocity of 1.3 m/s, the success rate of that system dropped sharply to 5 out of 15 flights. Furthermore, the LiDAR-based perception employed by \citet{karhunen2025fieldevaluationoptimizationlightweight} is known to suffer in open-space configurations (Table \ref{tab:combined_navigation} b) and in environments with foliage, where leaf-induced returns and sparse structure degrade reliability. The proposed vision-based system does not exhibit these limitations and maintains consistent performance across both dense and partially open forest environments.

By contrast, the vision-based approach of \citet{Karjalainen2023ARESULTS} is primarily limited by insufficient detection of thin obstacles such as small branches, resulting in substantially lower performance, with only 9 out of 19 successful flights in the baseline forest. Overall, the proposed method uniquely combines high success rates with higher flight velocities and robustness across varying forest structures.

Beyond the quantitative success rates, qualitative flight patterns further illustrate the robustness of the system. As observed in Figure \ref{fig:all_paths_c}, the drone occasionally executed loop-like maneuvers to escape local dead-end regions before continuing toward the goal. These behaviors indicate that the proposed navigation pipeline can reliably recover from constrained, ambiguous situations and reorient itself without external intervention, reinforcing its resilience in complex forest environments.

Despite these advances, overall reliability remains bounded by the quality of the depth representation. The metrics associated with the acceleration and smoothness values in the medium and difficult forest (Table \ref{tab:flight_results_set1} and \ref{tab:flight_results_set6}) are substantially lower than those recorded in the very difficult forest (Table \ref{tab:flight_results_set2}), reflecting the benefits of reduced clutter and more predictable free-space geometry, which allowed the planner to act more anticipatorily and less reactively. The proposed method reached a performance bottleneck in the most challenging forest environment (very difficult forest), where it achieved an 80 \% success rate, primarily due to depth-sensing limitations. Thin branches, twigs, and dense undergrowth continue to pose major perception challenges, as they frequently produce undefined or noisy disparity values that propagate through the planning pipeline. Although the depth improver significantly reduces local inconsistencies, it cannot reconstruct geometry that is fundamentally unobserved by the sensor.

These perceptual gaps occasionally lead to overly optimistic predictions from the planner, emphasizing the need for continued research into depth completion and multimodal perception, potentially combining RGB, LiDAR, or learned monocular priors to achieve more complete 3D understanding.


\subsection{Challanges and Future Work}

While the improved navigation system demonstrated strong real-world performance and a notable increase in robustness compared to earlier iterations, several challenges remain that constrain perception and generalization capabilities in complex forest environments.

The most persistent limitation lies in the quality of the depth representation used by the perception module. Although the proposed depth enhancement network significantly reduces sensor noise and partially restores missing structural information, its output remains inherently constrained by the input provided by the stereo camera. If certain structures, such as thin branches or foliage edges, are completely absent in the original depth image due to low reflectivity, lighting, or occlusions, the enhancement model cannot fully reconstruct them. Consequently, the drone may still underestimate the proximity of fine obstacles, especially in dense and heterogeneous vegetation, where small errors in depth can lead to unsafe proximity to branches. This limitation was particularly evident in experiments with the original SEVAE-ORACLE, which only completed 5 out of 15 flights in the very difficult forest at 1.0 m/s. Fine-tuning improved performance modestly (8/15 successful flights), but full reliability was only achieved by the proposed method (15/15 flights) even when increasing the target speed to 1.3 m/s. Future work will therefore focus on improving depth completion and reconstruction accuracy by incorporating complementary modalities, for example using RGB data, and by exploring transformer-based sparse-to-dense depth prediction architectures trained on real-world forest data.

A related challenge concerns the fusion of multiple visual cues. While the current system relies primarily on depth input, incorporating RGB information could substantially improve semantic understanding and help infer obstacle boundaries where geometric data is incomplete. A multimodal autoencoder that jointly learns from RGB and depth could improve generalization in low-light and high-clutter scenarios, producing more consistent representations for downstream navigation. This approach may help maintain the 100 \% success rate observed in medium-density forest experiments while further increasing robustness in dense or extremely cluttered environments.

Another avenue for improvement involves adaptive motion planning. The present planner operates using a discrete motion primitive library, which, while computationally efficient, limits system flexibility in complex environments. In dense forests, where obstacles vary in shape, spacing, and motion dynamics, for example swaying branches, future systems could benefit from learning-based or continuous planners that leverage uncertainty-aware predictions from the perception module. Superior performance of the proposed method in both 1.0 m/s and 1.3 m/s trials suggests that integrating richer predictive models could further enhance smoothness and reduce reactive maneuvers under high-speed navigation.

From a sensing perspective, improving reliability of 3D perception remains critical. Integrating lightweight alternative depth sensors, such as solid-state LiDARs or structured-light systems, could offer higher geometric accuracy, though at a cost in payload and power consumption. A practical intermediate step would be inclusion of a depth refinement or completion network, such as GuideNet \citep{tang2020learning}, to enhance stereo depth consistency before encoding, particularly in highly dense or occluded regions.

Finally, scaling the approach toward multi-agent or cooperative navigation \citep{tian2022kimera, lajoie2023swarm} under forest canopies represents a promising long-term direction. Sharing perception and localization data among multiple drones could compensate for individual sensor limitations and enable robust exploration in large-scale GNSS-denied environments, potentially improving overall success rates in highly challenging scenarios.

\section{Conclusion}

This work presented an enhanced vision-based autonomous navigation system for drones operating in dense forest environments. The system integrates a semantically-enhanced autoencoder for perception with a collision prediction network for decision-making, supported by a real-time supervisory safety layer. Through architectural refinements and targeted training on synthetic data, the system demonstrated significant improvement in interpreting depth information and navigating in highly cluttered, GNSS-denied conditions.

Extensive real-world experiments were conducted to validate system performance across three forest difficulty regimes: medium, high, and very high. In the medium forest, all 15 autonomous flights were successfully completed, demonstrating stable trajectory tracking and smooth motion without failures. In the difficult forest environment, considered as the baseline scenario, the proposed method achieved a 100\% success rate at both 1.0 m/s and 1.3 m/s target speeds. This level of performance was not matched by existing approaches. Among the compared systems, only the LiDAR-based method achieved a comparable success rate in the baseline forest, but only at a substantially lower average velocity, while its performance degraded sharply at higher speeds. Vision-based baselines, including SEVAE-ORACLE and its fine-tuned variant, exhibited significantly lower success rates under the same conditions. These results confirm that the combination of semantically enriched perception, predictive collision evaluation, and supervisory safety enables robust navigation even under extreme visual and structural complexity. In the very difficult forest, despite the increased clutter and reduced visibility, the system achieved an 80\% success rate, maintaining stable motion in the successful flights and demonstrating robust navigation under the most challenging conditions. 

On the other side, this last experiment further revealed the primary limitation of the current system, which is dependency on the fidelity of stereo depth input. While the depth enhancement module reduces sensor noise and fills partially missing regions, it cannot recover structures entirely absent from the original sensor data, such as thin branches or leaves. This limitation highlights the intrinsic constraints of stereo depth sensing in natural environments and motivates future research into multimodal perception, depth completion, and transformer-based sparse-to-dense prediction models.

Overall, the results demonstrate that simulation-trained, deep learning–based navigation systems can generalize effectively to complex real-world forest environments when complemented by task-specific perception refinements and robust safety-aware decision policies. The combination of learned depth encoding, predictive planning, and supervisory safety enables near-complete mission success across varying forest densities, supporting safe, adaptive behaviors such as velocity modulation and early avoidance maneuvers.

The successful completion of extensive field experiments under challenging environmental conditions provides strong evidence of system reliability and practical value for advanced remote sensing. These capabilities open new opportunities for high-resolution forest monitoring, environmental mapping, wildlife and habitat assessment, and rapid situational awareness in search-and-rescue operations. These findings establish a solid foundation for the next generation of autonomous aerial systems capable of safe, robust navigation in vegetation-rich, unstructured environments.

\section{Acknowledgement}

This research was funded by the Academy of Finland within
project “Learning techniques for autonomous drone based hyperspectral analysis of forest vegetation” (decision no. 357380). This study has been performed with affiliation to the Academy of Finland Flagship Forest–Human–Machine Interplay—Building Resilience, Redefining Value Networks and Enabling Meaningful Experiences (UNITE) (decision no. 357908).

\section{Declaration of generative AI and AI-assisted technologies in the manuscript preparation process}


During the preparation of this work, the author(s) used ChatGPT (OpenAI) for assistance in reviewing and refining the language of the manuscript. All content was critically reviewed and edited by the author(s), who take full responsibility for the final published version.




\bibliographystyle{elsarticle-harv} 
\bibliography{reference2}

\appendix
\section{Appendix}
\label{app1}
{
\renewcommand{\arraystretch}{1.0}

\begin{table*}[ht!]
\centering
\caption{flights data from the proposed method in the baseline difficult forest at 1 m/s.}
\label{tab:flight_results_set1}
\resizebox{\textwidth}{!}{%
\begin{tabular}{lcccccc}
\hline
\textbf{Flight} & \textbf{Disp. [m]} & \textbf{Avg. Vel. Disp. [m/s]} & \textbf{Avg. Odom. Vel. [m/s]} & \textbf{Avg. Accel. [m/s\textsuperscript{2}]} & \textbf{Smoothness\textsubscript{RMS} [m/s\textsuperscript{2}]} & \textbf{Successful} \\
\hline
Flight 1  & 58.98 & 0.88 & 0.96 & 0.45 & 0.66 & Yes \\
Flight 2  & 59.90 & 0.84 & 0.93 & 0.72 & 3.97 & Yes \\
Flight 3  & 59.72 & 0.79 & 0.87 & 0.68 & 0.91 & Yes \\
Flight 4  & 58.42 & 0.82 & 0.89 & 0.58 & 0.80 & Yes \\
Flight 5  & 58.19 & 0.82 & 0.91 & 0.62 & 0.83 & Yes \\
Flight 6  & 61.91 & 0.60 & 0.84 & 0.78 & 1.00 & Yes \\
Flight 7  & 58.64 & 0.77 & 0.87 & 1.30 & 17.85 & Yes \\
Flight 8  & 60.52 & 0.85 & 0.93 & 1.22 & 17.95 & Yes \\
Flight 9  & 58.34 & 0.87 & 0.92 & 0.57 & 0.78 & Yes \\
Flight 10 & 58.17 & 0.84 & 0.90 & 0.63 & 0.87 & Yes \\
Flight 11 & 57.10 & 0.88 & 0.93 & 0.57 & 0.82 & Yes \\
Flight 12 & 57.94 & 0.88 & 0.94 & 0.51 & 0.74 & Yes \\
Flight 13 & 58.39 & 0.74 & 0.83 & 0.66 & 0.88 & Yes \\
Flight 14 & 60.94 & 0.77 & 0.85 & 0.71 & 0.91 & Yes \\
Flight 15 & 58.51 & 0.92 & 0.97 & 0.46 & 0.66 & Yes \\
\hline
\textbf{Mean ± SD} 
& \textbf{59.05 ± 1.29} 
& \textbf{0.82 ± 0.08} 
& \textbf{0.90 ± 0.04} 
& \textbf{0.70 ± 0.25}  
& \textbf{3.31 ± 5.98}
& \textbf{100\%} \\
\hline
\end{tabular}
}
\end{table*}

\begin{table*}[ht!]
\centering
\caption{flights data from the proposed method in the baseline difficult forest at 1.3 m/s.}
\label{tab:flight_results_set3}
\resizebox{\textwidth}{!}{%
\begin{tabular}{lcccccc}
\hline
\textbf{Flight} & \textbf{Disp. [m]} & \textbf{Avg. Vel. Disp. [m/s]} & \textbf{Avg. Odom. Vel. [m/s]} & \textbf{Avg. Accel. [m/s\textsuperscript{2}]} & \textbf{Smoothness\textsubscript{RMS} [m/s\textsuperscript{2}]} & \textbf{Successful} \\
\hline
Flight 1  & 58.75 & 0.995 & 1.068 & 3.002 & 51.85 & Yes \\
Flight 2  & 58.96 & 1.082 & 1.174 & 0.790 & 1.054 & Yes \\
Flight 3  & 59.37 & 0.984 & 1.068 & 0.821 & 1.081 & Yes \\
Flight 4  & 58.28 & 0.748 & 0.899 & 3.249 & 48.68 & Yes \\
Flight 5  & 58.36 & 0.956 & 1.092 & 0.698 & 0.961 & Yes \\
Flight 6  & 58.21 & 0.930 & 1.032 & 0.922 & 1.185 & Yes \\
Flight 7  & 60.44 & 0.995 & 1.128 & 3.806 & 62.15 & Yes \\
Flight 8  & 59.82 & 1.054 & 1.165 & 0.927 & 4.642 & Yes \\
Flight 9  & 58.25 & 0.954 & 1.093 & 0.820 & 1.110 & Yes \\
Flight 10 & 58.24 & 1.038 & 1.138 & 0.836 & 1.156 & Yes \\
Flight 11 & 59.02 & 1.066 & 1.192 & 0.640 & 0.894 & Yes \\
Flight 12 & 58.76 & 1.017 & 1.141 & 0.844 & 1.113 & Yes \\
Flight 13 & 59.15 & 0.900 & 1.000 & 0.878 & 1.172 & Yes \\
Flight 14 & 59.68 & 0.997 & 1.079 & 1.016 & 1.314 & Yes \\
Flight 15 & 57.67 & 1.061 & 1.158 & 4.449 & 72.80 & Yes \\
\hline
\textbf{Mean ± SD} 
& \textbf{58.86 ± 0.74} 
& \textbf{0.985 ± 0.084} 
& \textbf{1.095 ± 0.077} 
& \textbf{1.580 ± 1.315}  
& \textbf{16.74 ± 26.79}
& \textbf{100\%} \\
\hline
\end{tabular}
}
\end{table*}

\begin{table*}[ht!]
\centering
\caption{flights data from fine-tuned seVAE-ORACLE \citep{DelColGuglielmo2024Autonomous} in the baseline difficult forest at 1 m/s.}
\label{tab:flight_results_set4}
\resizebox{\textwidth}{!}{%
\begin{tabular}{lcccccc}
\hline
\textbf{Flight} & \textbf{Disp. [m]} & \textbf{Avg. Vel. Disp. [m/s]} & \textbf{Avg. Odom. Vel. [m/s]} & \textbf{Avg. Accel. [m/s\textsuperscript{2}]} & \textbf{Smoothness\textsubscript{RMS} [m/s\textsuperscript{2}]} & \textbf{Successful} \\
\hline
Flight 1  & 58.06 & 0.981 & 1.009 & 0.953 & 11.79 & Yes \\
Flight 2  & 58.30 & 0.827 & 0.929 & 0.390 & 0.556 & Yes \\
Flight 3  & 58.65 & 0.828 & 0.946 & 1.083 & 13.43 & Yes \\
Flight 4  & 58.21 & 0.974 & 1.005 & 0.311 & 0.432 & Yes \\
Flight 5  & 36.22 & 0.972 & 1.019 & 1.382 & 20.15 & No \\
Flight 6  & 58.28 & 0.919 & 0.976 & 0.345 & 0.498 & Yes \\
Flight 7  & 27.22 & 0.977 & 1.025 & 0.349 & 0.446 & No \\
Flight 8  & 38.84 & 0.961 & 1.024 & 2.088 & 35.97 & No \\
Flight 9  & 58.27 & 0.826 & 0.941 & 1.031 & 15.83 & Yes \\
Flight 10 & 25.92 & 0.976 & 1.031 & 0.279 & 0.354 & No \\
Flight 11 & 28.99 & 0.933 & 1.014 & 0.327 & 0.427 & No \\
Flight 12 & 58.24 & 0.961 & 1.019 & 0.478 & 5.063 & Yes \\
Flight 13 & 5.50  & 1.042 & 1.040 & 0.323 & 0.485 & No \\
Flight 14 & 35.65 & 0.939 & 1.030 & 2.183 & 18.46 & No \\
Flight 15 & 58.20 & 0.943 & 1.011 & 2.414 & 40.59 & Yes \\
\hline
\textbf{Mean ± SD} 
& \textbf{44.30 ± 17.12} 
& \textbf{0.937 ± 0.064} 
& \textbf{1.001 ± 0.036} 
& \textbf{0.929 ± 0.759} 
& \textbf{10.97 ± 13.31} 
& \textbf{53\%} \\
\hline
\end{tabular}
}
\end{table*}

\begin{table*}[ht!]
\centering
\caption{flights data from seVAE-ORACLE \citep{Nguyen2024Uncertainty-awareNetworks} in the baseline difficult forest at 1 m/s.}
\label{tab:flight_results_set5}
\resizebox{\textwidth}{!}{%
\begin{tabular}{lcccccc}
\hline
\textbf{Flight} & \textbf{Disp. [m]} & \textbf{Avg. Vel. Disp. [m/s]} & \textbf{Avg. Odom. Vel. [m/s]} & \textbf{Avg. Accel. [m/s\textsuperscript{2}]} & \textbf{Smoothness\textsubscript{RMS} [m/s\textsuperscript{2}]} & \textbf{Successful} \\
\hline
Flight 1  & 53.36 & 0.90 & 1.02 & 0.35 & 0.58 & No \\
Flight 2  & 23.67 & 0.98 & 1.00 & 0.28 & 0.38 & No \\
Flight 3  & 58.19 & 0.90 & 1.02 & 0.28 & 0.38 & Yes \\
Flight 4  & 58.63 & 0.90 & 1.02 & 0.54 & 6.04 & Yes \\
Flight 5  & 41.50 & 0.95 & 1.01 & 0.49 & 2.56 & No \\
Flight 6  & 57.88 & 0.92 & 1.02 & 1.47 & 28.13 & Yes \\
Flight 7  & 34.18 & 1.00 & 1.01 & 0.67 & 3.27 & No \\
Flight 8  & 58.71 & 0.88 & 1.01 & 1.37 & 16.35 & Yes \\
Flight 9  & 58.39 & 0.95 & 1.03 & 0.34 & 0.48 & Yes \\
Flight 10 & 35.25 & 0.97 & 0.99 & 0.37 & 0.51 & No \\
Flight 11 & 29.35 & 0.99 & 1.00 & 0.43 & 0.61 & No \\
Flight 12 & 56.39 & 0.95 & 1.02 & 0.49 & 4.60 & No \\
Flight 13 & 14.57 & 1.01 & 1.01 & 1.09 & 9.48 & No \\
Flight 14 & 24.08 & 0.97 & 0.99 & 0.28 & 0.41 & No \\
Flight 15 & 24.16 & 1.00 & 1.01 & 0.30 & 0.47 & No \\
\hline
\textbf{Mean ± SD}
& \textbf{41.89 ± 16.19}
& \textbf{0.95 ± 0.04}
& \textbf{1.01 ± 0.01}
& \textbf{0.58 ± 0.40}
& \textbf{4.95 ± 7.83}
& \textbf{33\%} \\
\hline
\end{tabular}
}
\end{table*}

\begin{table*}[ht!]
\centering
\caption{flights data from the proposed method in the medium forest at 1 m/s.}
\label{tab:flight_results_set6}
\resizebox{\textwidth}{!}{%
\begin{tabular}{lcccccc}
\hline
\textbf{Flight} & \textbf{Disp. [m]} & \textbf{Avg. Vel. Disp. [m/s]} & \textbf{Avg. Odom. Vel. [m/s]} & \textbf{Avg. Accel. [m/s\textsuperscript{2}]} & \textbf{Smoothness\textsubscript{RMS} [m/s\textsuperscript{2}]} & \textbf{Successful} \\
\hline
Flight 1  & 58.46 & 0.93 & 0.95 & 0.35 & 0.60 & Yes \\
Flight 2  & 58.52 & 0.90 & 0.95 & 0.34 & 0.57 & Yes \\
Flight 3  & 59.03 & 0.83 & 0.94 & 0.78 & 8.76 & Yes \\
Flight 4  & 58.29 & 0.93 & 0.96 & 0.32 & 0.52 & Yes \\
Flight 5  & 58.14 & 0.96 & 1.00 & 0.23 & 0.38 & Yes \\
Flight 6  & 58.01 & 0.95 & 0.97 & 0.86 & 10.36 & Yes \\
Flight 7  & 58.38 & 0.98 & 0.98 & 0.24 & 0.38 & Yes \\
Flight 8  & 58.76 & 0.91 & 0.98 & 0.42 & 2.80 & Yes \\
Flight 9  & 58.57 & 0.99 & 0.99 & 0.34 & 2.33 & Yes \\
Flight 10 & 58.35 & 0.90 & 0.95 & 1.73 & 29.87 & Yes \\
Flight 11 & 58.17 & 0.89 & 0.93 & 0.44 & 0.70 & Yes \\
Flight 12 & 58.42 & 0.92 & 0.95 & 0.36 & 0.59 & Yes \\
Flight 13 & 58.49 & 0.93 & 0.93 & 0.38 & 0.63 & Yes \\
Flight 14 & 59.62 & 0.90 & 0.94 & 0.44 & 0.67 & Yes \\
Flight 15 & 58.41 & 0.93 & 0.94 & 0.58 & 5.56 & Yes \\
\hline
\textbf{Mean ± SD}
& \textbf{58.51 ± 0.40}
& \textbf{0.92 ± 0.04}
& \textbf{0.96 ± 0.02}
& \textbf{0.52 ± 0.38}
& \textbf{4.31 ± 7.76}
& \textbf{100\%} \\
\hline
\end{tabular}
}
\end{table*}

\begin{table*}[ht!]
\centering
\caption{flights data from the proposed method in the very difficult forest at 1 m/s}
\label{tab:flight_results_set2}
\resizebox{\textwidth}{!}{%
\begin{tabular}{lcccccc}
\hline
\textbf{Flight} & \textbf{Disp. [m]} & \textbf{Avg. Vel. Disp. [m/s]} & \textbf{Avg. Odom. Vel. [m/s]} & \textbf{Avg. Accel. [m/s\textsuperscript{2}]} & \textbf{Smoothness\textsubscript{RMS} [m/s\textsuperscript{2}]} & \textbf{Successful} \\
\hline
Flight 1  & 58.71 & 0.87 & 0.96 & 0.52 & 0.68 & Yes \\
Flight 2  & 58.23 & 0.84 & 0.95 & 0.56 & 0.76 & Yes \\
Flight 3  & 19.99 & 0.79 & 0.88 & 2.53 & 27.17 & No \\
Flight 4  & 58.46 & 0.59 & 0.79 & 0.65 & 0.89 & Yes \\
Flight 5  & 43.25 & 0.81 & 0.91 & 0.99 & 7.29 & No \\
Flight 6  & 59.63 & 0.73 & 0.88 & 0.84 & 1.04 & Yes \\
Flight 7  & 58.49 & 0.85 & 0.92 & 1.38 & 22.06 & Yes \\
Flight 8  & 58.48 & 0.82 & 0.91 & 0.78 & 0.97 & Yes \\
Flight 9  & 58.74 & 0.53 & 0.82 & 1.41 & 18.25 & Yes \\
Flight 10 & 57.51 & 0.79 & 0.90 & 1.03 & 7.81 & Yes \\
Flight 11 & 60.49 & 0.85 & 0.94 & 0.66 & 0.84 & Yes \\
Flight 12 & 17.05 & 0.91 & 0.92 & 0.60 & 0.77 & No \\
Flight 13 & 58.37 & 0.83 & 0.92 & 0.68 & 0.88 & Yes \\
Flight 14 & 58.82 & 0.80 & 0.95 & 2.66 & 54.47 & Yes \\
Flight 15 & 58.82 & 0.74 & 0.89 & 0.73 & 0.93 & Yes \\
\hline
\textbf{Mean ± SD} 
& \textbf{52.34 ± 14.32} 
& \textbf{0.78 ± 0.10} 
& \textbf{0.90 ± 0.05} 
& \textbf{1.07 ± 0.68}  
& \textbf{9.65 ± 15.23}
& \textbf{80\%} \\
\hline
\end{tabular}
}
\end{table*}

}







\end{document}